\pdfoutput=1

\documentclass[11pt]{article}
\usepackage[table,dvipsnames,x11names]{xcolor}

\usepackage[final]{acl}

\usepackage{enumitem}
\usepackage[most]{tcolorbox}
\usepackage{lipsum}  
\usepackage{tikz}
\usepackage{tcolorbox}
\usepackage{pifont}
\usepackage{float}

\usepackage{times}
\usepackage{latexsym}
\usepackage{booktabs}
\usepackage{amssymb}  
\usepackage{amsmath}  
\usepackage{booktabs}  
\usepackage{multirow}  
\definecolor{darkpurple}{RGB}{102, 0, 153}
\usepackage{hyperref}

\usepackage[ruled,vlined,linesnumbered]{algorithm2e}

\usepackage{tabularx}
\usepackage{array, booktabs}

\makeatletter
\def\@fnsymbol#1{\ensuremath{\ifcase#1\or *\or \dagger\or \ddagger\or
   \mathsection\or \mathparagraph\or \|\or **\or \dagger\dagger
   \or \ddagger\ddagger \else\@ctrerr\fi}}
\makeatother

\definecolor{softblue}{RGB}{65, 157, 250}
\usepackage{hyperref}  
\hypersetup{
    colorlinks,
    linkcolor=darkpurple,
    anchorcolor=blue,
    citecolor=softblue
}
\usepackage[T1]{fontenc}

\usepackage[utf8]{inputenc}

\usepackage{microtype}

\usepackage{inconsolata}

\usepackage{graphicx}

\usepackage{tcolorbox}
\definecolor{lightred}{RGB}{230, 180, 180}
\definecolor{lightgreen}{RGB}{200, 225, 200}
\definecolor{softred}{RGB}{180, 50, 50}
\definecolor{softgreen}{RGB}{30, 120, 60}

\usepackage{siunitx}
\title{DeepSieve: Information Sieving via LLM-as-a-Knowledge-Router}

\author{
Minghao Guo$^1$\quad
Qingcheng Zeng$^2$\quad
Xujiang Zhao$^3$\quad
Yanchi Liu$^3$\quad \\
\textbf{Wenchao Yu}$^3$\quad 
\textbf{Mengnan Du}$^4$\quad 
\textbf{Haifeng Chen}$^3$\quad 
\textbf{Wei Cheng}$^3$\thanks{Corresponding Author Email: weicheng@nec-labs.com} \\
  $^1$Rutgers University\;\;\; $^2$Northwestern University\;\;\; $^3$NEC Laboratories America\;\;\;    $^4$NJIT\;\;\;
}

\begin{document}
\maketitle
\begin{abstract}
Large Language Models (LLMs) excel at many reasoning tasks but struggle with knowledge-intensive queries due to their inability to dynamically access up-to-date or domain-specific information. Retrieval-Augmented Generation (RAG) has emerged as a promising solution, enabling LLMs to ground their responses in external sources. However, existing RAG methods lack fine-grained control over both the query and source sides, resulting in noisy retrieval, shallow reasoning, and limited adaptability to heterogeneous knowledge sources. In this work, we introduce \textit{\textbf{DeepSieve}}, a novel RAG method that incorporates information sieving via LLM-as-a-knowledge-router. DeepSieve breaks down complex queries into structured sub-queries and recursively routes each to the most appropriate knowledge source, filtering out irrelevant information through a multi-stage information sieving process. This modular and transparent approach ensures that DeepSieve remains adaptable across diverse information needs. Experiments on three multi-hop QA benchmarks involving heterogeneous sources show that DeepSieve achieves greater reasoning depth, retrieval precision, and interpretability compared to conventional RAG approaches. Our codes are available at \href{https://github.com/MinghoKwok/DeepSieve}{https://github.com/MinghoKwok/DeepSieve}.
\end{abstract}

\section{Introduction}

Large language models (LLMs) have achieved remarkable progress across a wide range of natural language tasks and have demonstrated strong reasoning abilities, including in domains such as mathematics~\cite{guo2025deepseek}, fake news detection \cite{yi2025challengesinnovationsllmpoweredfake, jin2024exploring}, and commonsense reasoning~\cite{yao2023react}. Yet these LLMs often falter when facing knowledge-intensive questions that require up-to-date or domain-specific information~\cite{guu2020retrieval, asai2023self, fan2024survey}. This limitation stems from the fixed nature of LLM parameters, which prevents dynamic access to external knowledge and frequently results in hallucinations or factually inaccurate outputs~\cite{huang2025survey}. Retrieval-Augmented Generation (RAG) has emerged as a powerful paradigm for equipping LLMs with access to external knowledge~\cite{lewis2020retrieval, citation-0,10.1145/3726302.3729957}, enabling them to tackle complex information-seeking tasks more effectively. Recent advances such as GraphRAG, HippoRAG, and so on~\cite {hippo, edge2024local, zhang2025survey} demonstrate the promise of structured and memory-augmented retrieval in improving factual accuracy and multi-hop reasoning.

\begin{figure}[t]
  \centering
  \includegraphics[width=\linewidth]{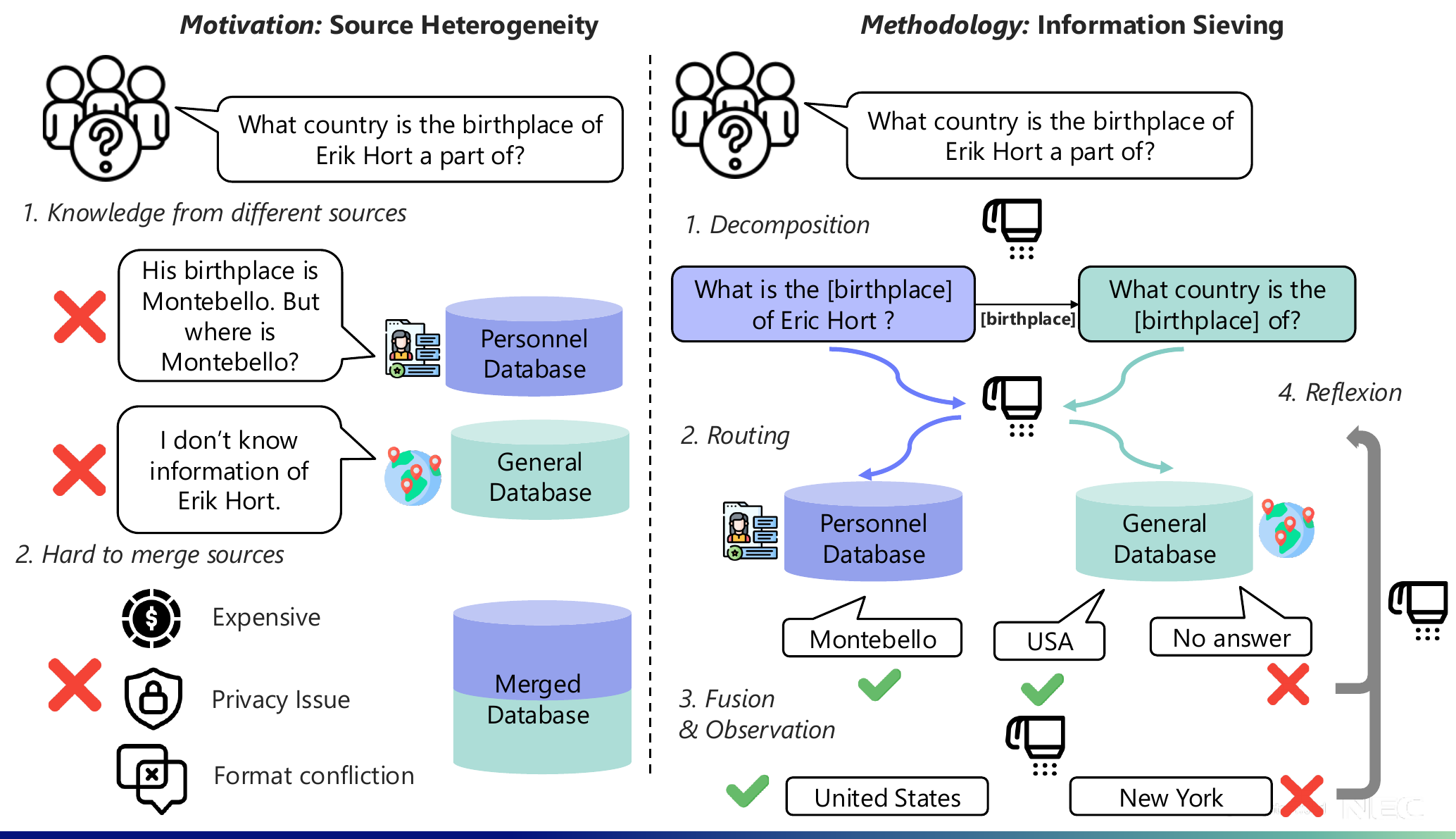}
\caption{
  \textbf{Motivation and Overview: } 
  Left: Compositional queries are hard to answer under source heterogeneity (e.g., structured, private, and unmergeable databases). 
  Right: DeepSieve performs decomposition, source-aware routing, and iterative fusion to enable structured reasoning.
}

  \vspace{-1em}
  \label{fig:overview}
\end{figure}

However, existing RAG systems still struggle with a fundamental limitation: the lack of an effective \textit{information sieving} mechanism across the reasoning pipeline. In practice, this deficiency manifests in two critical forms: \textit{query-side sieving} and \textit{source-side sieving}. On the query side, most systems treat user queries as atomic units and directly retrieve information without decomposing or analyzing their underlying semantic structure~\cite{gao2022interleaved, yao2023react, yin2023s}. This prevents the model from isolating key subgoals or reasoning steps, which is essential for multi-hop or compositional question answering.

On the source side, knowledge can be heterogeneous both across different sources (e.g., unstructured corpora, structured APIs, private databases) and within a single source that contains multiple domains or topics. However, most existing RAG systems retrieve from a flat, unified index without considering differences in domain, format, access modality, or granularity~\cite{chen2017reading, izacard2020leveraging, gao2022pal}. This often results in mismatched content, irrelevant retrievals, and unnecessary computational cost. Moreover, many knowledge sources cannot be merged into a single retrieval index because of privacy constraints, structural incompatibility, or deployment considerations~\cite{nguyen2023semix, xu2025amemagenticmemoryllm}, further highlighting the need for selective and modular retrieval strategies.

To bridge this knowledge gap, 
we propose \textbf{\textit{DeepSieve}}, a framework that performs retrieval-augmented reasoning through multi-stage information sieving. As shown in the Figure \ref{fig:overview}, DeepSieve consists of four key components: \textit{question decomposition}, \textit{thought generation}, \textit{data-source routing}, and \textit{recursive reflexion}. The pipeline begins by decomposing a complex input query into a set of structured sub-questions using an LLM-based planner~\cite{yin2023s, trivedi2022musique}. Each sub-question is then independently processed to generate a “thought”, a latent reasoning step that guides retrieval planning~\cite{yao2023react, zhang2023rewoo}. Based on this thought, DeepSieve identifies the most appropriate source (e.g., API, SQL, RAG corpus) and generates a corresponding action plan. This source-aware routing builds on the idea of using LLMs as controllers for modular tool access~\cite{schick2023toolformer, xu2025amemagenticmemoryllm}.

Once evidence is retrieved, the system evaluates whether the information sufficiently addresses the sub-question. If the answer is invalid, DeepSieve enters a reflection loop to reassess the thought, revise the action plan, or reselect the source, iterating until resolution or timeout~\cite{madaan2023selfrefine, baek-etal-2025-probing}. Finally, after all sub-questions are addressed, DeepSieve aggregates the answers into a coherent output using LLM module, which also generates intermediate rationales~\cite{wei2022chain, creswell2022selection}.

This iterative, modular design allows DeepSieve to sieve and integrate knowledge progressively, making it particularly suited for multi-hop reasoning and heterogeneous information access. Empirically, DeepSieve consistently outperforms the baselines across all three benchmarks, achieving average F1/EM scores of 58.9/49.3 with DeepSeek-V3 and 51.2/41.4 with GPT-4o, surpassing both RAG (e.g., HippoRAG~\cite{hippo}) and agentic (e.g., ReAct~\cite{yao2023react}) baselines, while using significantly fewer tokens. To sum up, our paper makes the following key contributions:

\begin{itemize}
    \item We identify the structural and semantic heterogeneity of real-world knowledge sources as a core challenge in RAG. To address this, we propose the \textbf{DeepSieve} framework, which introduces \textit{information sieving} and, for the first time, uses an \textbf{LLM-as-a-knowledge-router} to dynamically decompose queries and dispatch sub-questions to heterogeneous sources.
    \item Our experiments demonstrate that DeepSieve not only satisfies multi-source settings but also improves performance in standard single-source scenarios. Even when operating over a unified corpus, our system yields better retrieval precision and answer accuracy.
    \item We design DeepSieve as a modular and extensible method. It supports plug-and-play integration with diverse tools, retrieval backends, and RAG models. It provides a flexible backbone for future RAG architectures.
\end{itemize}

\section{Methodology}

\begin{figure*}[t]
  \centering
  \includegraphics[width=0.95\linewidth]{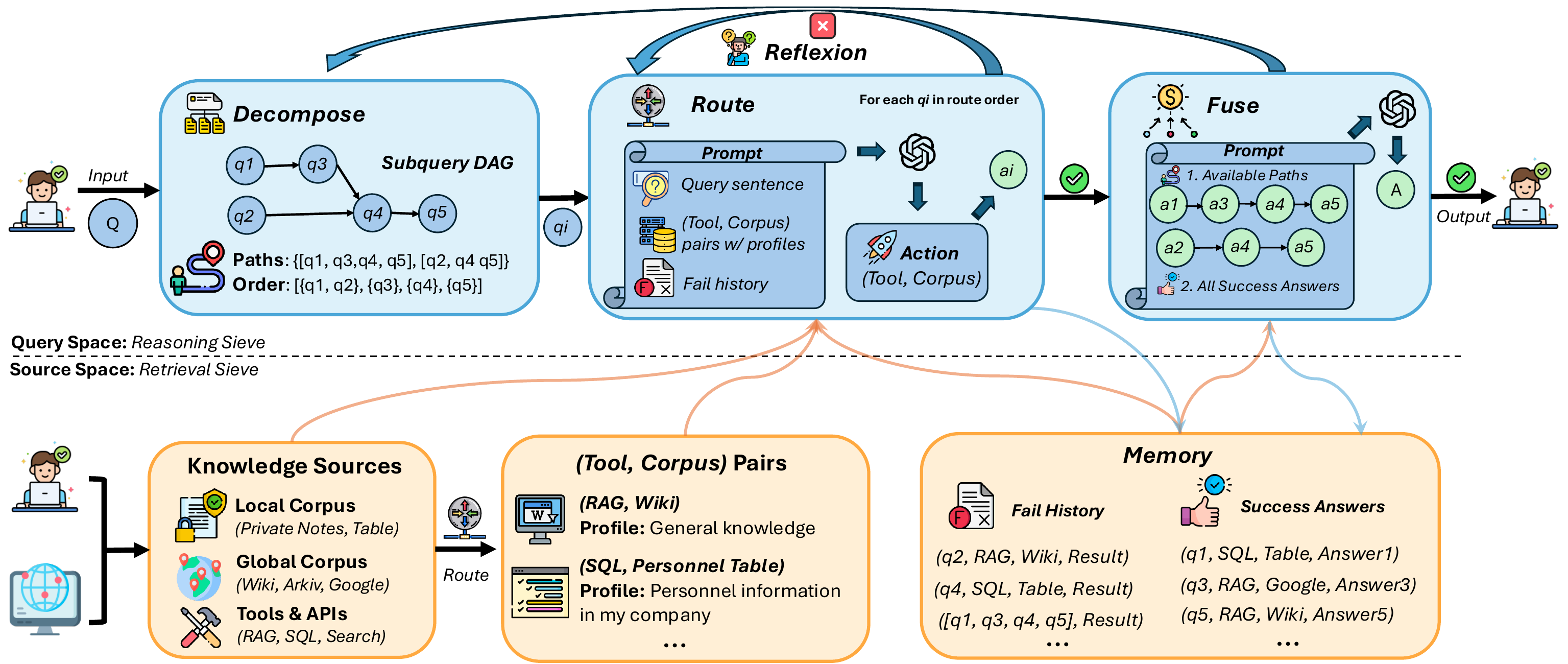}
  \caption{\textbf{\textit{DeepSieve workflow}} with multi-step reasoning: A complex query is first decomposed into diverse subqueries, like \textit{a directed acyclic graph (DAG)}. For each path in the DAG, an LLM generates a plan to select knowledge sources via routing. Failed retrievals will trigger re-routing or re-decomposition in the workflow. Retrieved subanswers are stored in memory and later fused across paths to form a final answer. }
 \label{fig:pipeline}
\end{figure*}

We propose \textbf{DeepSieve}, a novel RAG method that performs information sieving across both query and source spaces. DeepSieve addresses the challenge of reasoning over heterogeneous knowledge sources, spanning diverse formats, access modalities, and domains, by employing a layered process of decomposition, routing, reflection, and fusion.

As shown in Figure~\ref{fig:overview} and Algorithm~\ref{alg:deepsieve}, DeepSieve decomposes input queries into subqueries, routes each to an appropriate \textit{(Tool, Corpus)} pair, iteratively refines insufficient answers via reflexion, and ultimately fuses resolved subanswers into a coherent final response.

\paragraph{Notation.} Let $Q$ denote the original input query; let $\mathcal{S} = \{(T_k, C_k)\}$ denote the source set, where $T_k$ is a tool and $C_k$ is a corpus. DeepSieve performs the following modular operations:

\noindent
$\texttt{Decompose}(Q) \rightarrow \{q_i\}_{i=1}^n$: Decomposes the query into structured subqueries. \\
$\texttt{Route}(q_i, \mathcal{S}) \rightarrow s_i$: Selects a knowledge source $s_i = (T_i, C_i)$ from the source set $\mathcal{S}$. \\
$\texttt{Retrieve}(q_i, s_i) \rightarrow a_i$: Retrieves an answer $a_i$ from source $s_i$ for subquery $q_i$. \\
$\texttt{Reflect}(q_i, a_i) \rightarrow IF\_REPLAN$: If $a_i$ is insufficient, determines whether to replan $q_i$. \\
$\texttt{Fuse}(\{a_i\}) \rightarrow \hat{A}$: Aggregates all valid subanswers into a final response.

\paragraph{System Input and Execution.}
Given a query \( Q \) and a source pool \( \mathcal{S} = \{(T_k, C_k)\} \) of tool–corpus pairs, DeepSieve process the query in 4 stages as shown in Algorithm~\ref{alg:deepsieve} and illustrated from Section~\ref{sec:decomposition} to Section~\ref{sec:fusion}.



\begin{algorithm}[t]
\caption{DeepSieve Pipeline}
\label{alg:deepsieve}
\KwIn{Natural language query $Q$; source set $\mathcal{S}$}
\KwOut{Final answer $\hat{A}$}

$\{q_1, \ldots, q_n\} \leftarrow \texttt{Decompose}(Q)$ 

\ForEach{$q_i$ in execution order}{
    $(T_i, C_i) \leftarrow \texttt{Route}(q_i, \mathcal{S})$ 
    $a_i \leftarrow \texttt{Retrieve}(q_i, T_i, C_i)$ 


    $\texttt{IF\_REPLAN} \leftarrow \texttt{Reflect}(q_i, a_i)$ \;
    \If{$\texttt{IF\_REPLAN}$}{
        $(T_i, C_i) \leftarrow \texttt{Route}(q_i, \mathcal{S} \setminus \{(T_i, C_i)\})$ \;
        $a_i \leftarrow \texttt{Retrieve}(q_i, T_i, C_i)$ \;
    }

    Store $(q_i, (T_i, C_i), a_i)$ into memory $\mathcal{M}$ \;
}

$\hat{A} \leftarrow \texttt{Fuse}(\{a_i\}_{i=1}^n, \mathcal{M})$ 
\Return $\hat{A}$
\end{algorithm}

\subsection{Stage I: Query Decomposition}
\label{sec:decomposition}

Given a complex query \(\mathcal{Q}\), DeepSieve first invokes an LLM-based planner to decompose it into a set of structured subquestions \(\{q_1, q_2, \dots, q_n\}\). This step acts as a \textit{semantic sieve}, transforming monolithic input into a directed acyclic graph (DAG) of subgoals. Each node in the DAG represents an atomic reasoning unit, while edges capture dependency constraints resolved at fusion time. 


\subsection{Stage II: Knowledge Routing via LLM-as-Router}
\label{sec:routing}

For each subquestion \(q_i\), the system employs an LLM-based router to select a tool–corpus pair \((T_i, C_i)\) from the source pool \(\mathcal{S} = \{(T_k, C_k)\}\). This selection is guided by a structured routing prompt that encodes (i) the semantics of \(q_i\), (ii) the profile metadata of each source (e.g., domain, format, privacy level), and (iii) a fail history \(\mathcal{M}_\text{fail}\) of previous retrieval attempts.

The router returns a source \(s_i = (T_i, C_i)\), and the system invokes the retrieval tool \(T_i\) on corpus \(C_i\) to obtain an answer candidate \(a_i\). 

\subsection{Stage III: Observation and Reflexion}

If the retrieved answer \(a_i\) is deemed unsatisfactory, e.g., incomplete, irrelevant, or ambiguous, DeepSieve triggers a reflexion step to reassess the current subquery \(q_i\). Instead of modifying the subquery content, the system re-routes \(q_i\) to select an alternative tool–corpus pair \(s_i' = (T_i', C_i')\), and attempts retrieval again.

This process is guided by the memory module \(\mathcal{M}\), which records all attempted subqueries and outcomes. Specifically, failed retrievals are stored in \(\mathcal{M}_\text{fail}\) as tuples \((q_i, s_i, \texttt{Result})\), helping the router avoid redundant sources in future attempts. Successful results are stored in \(\mathcal{M}_\text{succ}\) as \((q_i, s_i, a_i)\), forming trusted evidence for final answer fusion.

\subsection{Stage IV: Answer Fusion}
\label{sec:fusion}

After all subquestions have been individually resolved, DeepSieve invokes a fusion module to perform final aggregation. This module collects the set of successful subanswers \(\{a_i\}_{i=1}^{n}\) from memory \(\mathcal{M}_\text{succ}\), and synthesizes them into a globally coherent answer \(\hat{A}\).

The fusion process leverages the DAG structure defined during query decomposition, which encodes both the reasoning order and dependency relationships among subquestions. It considers all valid reasoning paths that traverse the subquestion graph and selects consistent subanswers along these paths for inclusion. In cases where conflicting evidence is encountered, DeepSieve optionally performs global inference using an LLM to resolve contradictions and generate a unified response.


\subsection{Modularity and Extensibility}

DeepSieve is designed with a modular architecture that supports seamless integration of heterogeneous tools and knowledge sources. Each core component, decomposition, routing, retrieval, reflexion, and fusion, can be independently replaced or extended without modifying the overall control flow. Knowledge sources are abstracted as \texttt{(Tool, Corpus)} pairs, each annotated with a natural language profile that guides source selection during routing. This abstraction enables plug-and-play extension: adding a new retriever (e.g., \texttt{BM25}, \texttt{FAISS}, \texttt{ColBERTv2}) or a new source (e.g., SQL, API) only requires registering its wrapper and profile. The system also scales naturally to multi-source settings through semantic clustering or source-specific wrappers, eliminating the need for index merging or schema unification.


\section{Experiments}

We evaluate DeepSieve on multi-hop QA benchmarks to answer four core research questions:

\begin{itemize}
    \item \textbf{RQ1:} Does DeepSieve outperform traditional RAG baselines?
    \item \textbf{RQ2:} Is DeepSieve more efficient in inference cost than other agentic RAG methods?
    \item \textbf{RQ3:} How do decomposition, routing, and reflexion contribute to overall performances, respectively?
    \item \textbf{RQ4:} Can DeepSieve adapt flexibly across different retrievers and modular knowledge source configurations?

\end{itemize}

\subsection{Experimental Setup}

\begin{figure}[t]
\centering
\includegraphics[width=0.95\linewidth]{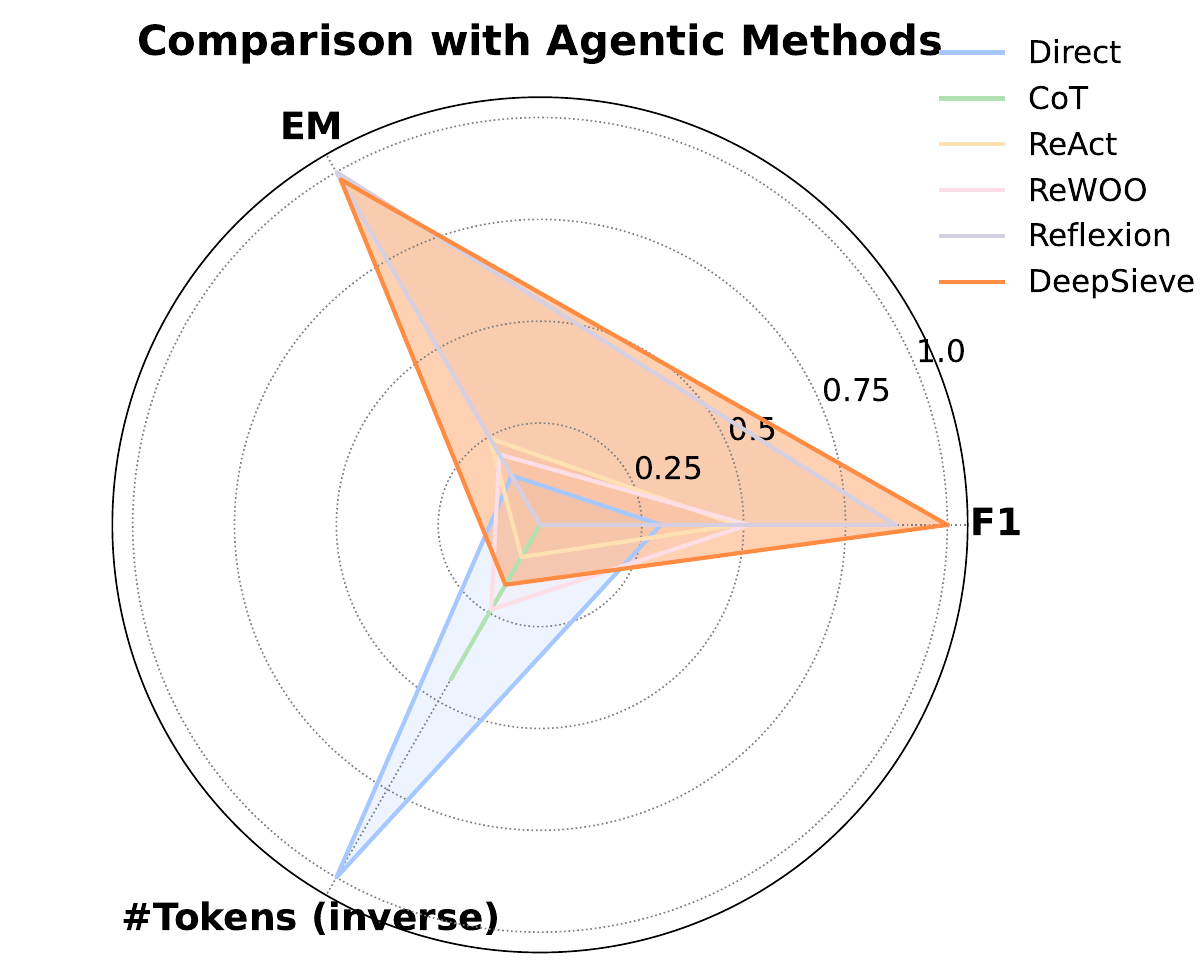}
\caption{Normalized radar plot comparing agentic methods based on their average scores across all benchmarks. The plot evaluates methods across three dimensions: F1 score, EM score, and token efficiency represented as \#Tokens (inverse). All metrics are normalized, and a higher value indicates better performance for each axis. A larger enclosed area signifies a superior trade-off between accuracy and computational cost.}
\label{fig:agentic_rag_radar}
\end{figure}

We evaluate on 3 benchmarks: MuSiQue~\cite{trivedi2022musique}, 2WikiMultiHopQA~\cite{ho-etal-2020-constructing}, and HotpotQA~\cite{yang2018hotpotqa}, following IRCoT~\cite{gao2022interleaved} to construct retrieval corpora with both supporting and distractor passages (1,000 dev questions). We use DeepSeek-V3~\cite{guo2025deepseek} and GPT-4o~\cite{openai2024gpt4o} as the backbone LLM. To simulate source heterogeneity, we partition each dataset into \textit{local} and \textit{global} segments using LLM-based profiles (see Appendix~\ref{appendix:dataset}). DeepSieve performs subquestion-level routing over these two source modules, allowing modular reasoning across simulated access boundaries. Since existing baselines do not support multi-source retrieval or dynamic source selection, we evaluate them using the original corpus, which is the same as the combination of local and global segments mentioned above, to ensure fair comparison. This setup allows us to isolate the impact of modular routing while maintaining compatibility across methods.

\paragraph{Baselines.} We compare against IRCoT~\cite{gao2022interleaved}, ColBERTv2~\cite{santhanam2022colbertv2},  HippoRAG~\cite{hippo}, and RAPTOR~\cite{sarthi2024raptor} as representative RAG paradigms baseline. We also include ReAct~\cite{yao2023react}, ReWOO~\cite{xu2023rewoo}, Reflexion~\cite{shinn2023reflexion}, and Chain-of-Thought (CoT)~\cite{wei2022chain} as reasoning and agentic baselines because DeepSieve utilizes not only RAG algorithm. The details of these baselines and their setups are listed in Appendix~\ref{appendix:baseline_details}.

\paragraph{Metrics.} 
We report Exact Match (EM) and F1 scores to evaluate answer correctness, where EM measures exact string match and F1 accounts for token-level overlap. To assess inference cost, we track the total number of tokens generated by the LLM across all reasoning steps.

\subsection{Main Performance Comparison (RQ1, RQ4)}

\begin{table*}[t]
\small
\centering
\caption{
Exact Match (EM) and F1 scores on MuSiQue, 2WikiMultihopQA, and HotpotQA under DeepSeek-V3 and GPT-4o. 
DeepSieve is evaluated using a simulated heterogeneous setup with local/global source partitioning; baselines use the merged corpus. 
Bold: best; ↑: improvement over best non-DeepSieve; superscripts indicate F1/EM gains.
}
\label{tab:main_qa_comparison}
\setlength{\tabcolsep}{4.5pt} 
\renewcommand{\arraystretch}{1.15} 

\newcolumntype{N}{S[
    table-format=2.1,
    table-space-text-post={\textsuperscript{+16.3\textuparrow}},
    mode=text,
    detect-weight=true
]}

\resizebox{\textwidth}{!}{%
\begin{tabular}{ l l *{8}{N} }
\toprule
& \textbf{Retriever} & \multicolumn{2}{c}{\textbf{MuSiQue}} & \multicolumn{2}{c}{\textbf{2Wiki}} & \multicolumn{2}{c}{\textbf{HotpotQA}} & \multicolumn{2}{c}{\textbf{Average}} \\
\cmidrule(lr){3-4} \cmidrule(lr){5-6} \cmidrule(lr){7-8} \cmidrule(lr){9-10}
& & {\makebox[0pt]{EM}} & {\makebox[0pt]{F1}} & {\makebox[0pt]{EM}} & {\makebox[0pt]{F1}} & {\makebox[0pt]{EM}} & {\makebox[0pt]{F1}} & {\makebox[0pt]{EM}} & {\makebox[0pt]{F1}} \\

\midrule
\multirow{8}{*}{\rotatebox{90}{\textbf{DeepSeek-V3}}}
& Naive RAG (MiniLM) & 20.5 & 26.1 & 26.4 & 31.3 & 30.4 & 42.8 & 25.8 & 33.4 \\
& ColBERTv2          & 17.1 & 27.2 & 32.9 & 43.8 & 42.9 & 57.2 & 31.0 & 42.7 \\
& HippoRAG           & 19.4 & 27.9 & 44.9 & 58.9 & 40.8 & 55.3 & 35.0 & 47.4 \\
& IRCoT + HippoRAG   & 21.4 & 33.4 & 46.5 & 63.1 & 45.1 & 58.9 & 37.7 & 51.8 \\
& GraphRAG           & 7.7  & 9.7  & 30.1 & 33.2 & 45.9 & 55.2 & 27.9 & 32.7 \\
& RAPTOR             & 18.2 & 28.9 & 38.6 & 52.1 & 52.9 & 66.5 & 37.5 & 50.2 \\
& \textbf{DeepSieve (Naive RAG)} & \textbf{36.0}\textsuperscript{+14.6\textuparrow} & \textbf{46.8}\textsuperscript{+13.4\textuparrow} & \textbf{62.8}\textsuperscript{+16.3\textuparrow} & \textbf{68.4}\textsuperscript{+5.3\textuparrow} & 49.0 & 61.6 & \textbf{49.3}\textsuperscript{+11.6\textuparrow} & \textbf{58.9}\textsuperscript{+7.1\textuparrow} \\
& \textbf{DeepSieve (GraphRAG)} & 30.0\textsuperscript{+8.6\textuparrow} & 36.6\textsuperscript{+3.2\textuparrow} & 49.2\textsuperscript{+2.7\textuparrow} & 53.8 & 49.4 & 61.6 & 42.9\textsuperscript{+5.2\textuparrow} & 50.7 \\
\midrule
\multirow{8}{*}{\rotatebox{90}{\textbf{GPT-4o}}}
& Naive RAG (MiniLM) & 19.4 & 27.1 & 26.7 & 29.8 & 29.8 & 44.5 & 25.3 & 33.8 \\
& ColBERTv2          & 15.5 & 26.4 & 33.4 & 43.3 & 43.4 & 57.7 & 30.8 & 42.5 \\
& HippoRAG           & 19.2 & 29.8 & 46.6 & 59.5 & 41.8 & 55.0 & 35.9 & 48.1 \\
& IRCoT + HippoRAG   & 21.9 & 33.3 & 47.7 & 62.7 & 45.7 & 56.2 & 38.4 & 50.7 \\
& GraphRAG           & 8.7  & 10.9 & 29.8 & 33.3 & 44.3 & 52.2 & 25.6 & 31.0 \\
& RAPTOR             & 18.8 & 29.7 & 39.3 & 51.6 & 52.4 & 66.3 & 36.8 & 49.2 \\
& \textbf{DeepSieve (Naive RAG)} & \textbf{26.7}\textsuperscript{+4.8\textuparrow} & \textbf{36.6}\textsuperscript{+3.3\textuparrow} & \textbf{48.3}\textsuperscript{+0.6\textuparrow} & 55.2 & 49.3 & 61.7 & \textbf{41.4}\textsuperscript{+3.0\textuparrow} & \textbf{51.2}\textsuperscript{+0.5\textuparrow} \\
& \textbf{DeepSieve (GraphRAG)} & 20.3 & 31.1 & 27.5 & 38.9 & 44.6 & 53.5 & 32.7 & 43.2 \\
\bottomrule
\end{tabular}
} %

\end{table*}


In this section, we aim to evaluate whether DeepSieve improves answer accuracy across multiple multi-hop QA datasets, and how it compares with both pure RAG and reasoning-based baselines.
Table~\ref{tab:main_qa_comparison} includes two DeepSieve variants to demonstrate its adaptability across retrieval paradigms: \textbf{Naive RAG}, which retrieves from a flat corpus using all-MiniLM-L6-v2, and \textbf{GraphRAG}, which builds on structure-aware retrieval via document-level links following the GraphRAG setting. Both variants use DeepSeek-V3 and GPT-4o as the backbone LLM with the identical decoding parameters, which illustrates DeepSieve’s modularity across different retrieval methods.

DeepSieve(Naive RAG) achieves the best F1 score on MuSiQue (46.8, +13.5 over IRCoT+HippoRAG) and 2WikiMultiHopQA (68.4, +5.3), highlighting the benefit of structured decomposition and source-aware retrieval. On HotpotQA, DeepSieve performs below RAPTOR, which benefits from the design of HotpotQA itself that favors models with entity linking and graph construction. However, unlike RAPTOR, DeepSieve performs better on average and operates in a fully online and modular manner, without requiring any static graph preprocessing or clustering. On average, DeepSieve (Naive RAG) achieves an F1 of 58.9, outperforming all baselines.

Under the GPT-4o setting, DeepSieve (Naive RAG) achieves an F1 of 61.7 on HotpotQA (Table~\ref{tab:agentic_comparison}), outperforming other multi-hop reasoning frameworks such as Chain-of-Thought (30.8), ReAct (39.6), and ReWOO (40.1). It also approaches the performance of Reflexion (46.7 vs. 49.3), while employing a more modular design with explicit decomposition and source-specific coordination.

\begin{tcolorbox}[
  colback=blue!5!white,
  colframe=black,
  coltitle=white,
  title=Takeaway: RQ1,
  fonttitle=\bfseries,
  sharp corners=southwest,
  enhanced,
  attach boxed title to top center={
    yshift=-2mm
  },
  boxed title style={
    colback=black,
    size=small,
    boxrule=0pt,
    rounded corners=southeast,
    sharp corners=north
  }
]
DeepSieve achieves the best average F1 across all datasets, outperforming both pure RAG baselines and agentic RAG method baselines without relying on static graphs.
\end{tcolorbox}

\subsection{Efficiency Comparison with Reasoning and Agentic Methods (RQ1, RQ2)}

\begin{table}[t]
\centering
\caption{Comparison of Reasoning \& Agentic RAG paradigms on \textbf{HotpotQA} using GPT-4o in terms of F1, EM, and token usage.}
\label{tab:agentic_comparison}
\begin{tabular}{lccc}
\toprule
\textbf{Paradigm} & \textbf{F1} & \textbf{EM} & \textbf{\#Tokens} \\
\midrule
Direct & 36.2 & 28.0 & \textbf{55.5} \\
CoT & 30.8 & 22.4 & 481.9 \\
ReAct & 39.6 & 32.2 & 9795.1 \\
ReWOO & 40.1 & 30.4 & 1986.2 \\
Reflexion & \textbf{62.5} & 46.7 & 37893.0 \\
DeepSieve & 61.7 & \textbf{49.3} & 3926.6 \\
\bottomrule
\end{tabular}
\end{table}

We then evaluate whether DeepSieve achieves its performance gains efficiently by comparing LLM token usage with other LLM-based reasoning and agentic methods. As shown in Table~\ref{tab:agentic_comparison}, DeepSieve attains higher accuracy while using significantly fewer tokens. On HotpotQA, DeepSieve achieves the highest EM score (49.3) and F1 (61.7), with an average of only 3.9K tokens per query, compared to Reflexion (37.9K) and ReAct (9.8K).

To better illustrate the performance–efficiency trade-off, Figure~\ref{fig:agentic_rag_radar} shows a normalized radar plot across three dimensions: F1, EM, and inverse token usage. DeepSieve covers the largest area, demonstrating strong overall performance across all metrics. While ReAct and Reflexion achieve similar accuracy, they require far more tokens. In contrast, Direct and CoT are efficient but lag in accuracy. By balancing all three dimensions, DeepSieve stands out as a cost-effective LLM-based RAG system.

\begin{tcolorbox}[
  colback=blue!5!white,
  colframe=black,
  coltitle=white,
  title=Takeaway: RQ2,
  fonttitle=\bfseries,
  sharp corners=southwest,
  enhanced,
  attach boxed title to top center={
    yshift=-2mm
  },
  boxed title style={
    colback=black,
    size=small,
    boxrule=0pt,
    rounded corners=southeast,
    sharp corners=north
  }
]
DeepSieve uses significantly fewer tokens than other LLM-based systems while achieving higher accuracy, showing strong cost-effectiveness.

\end{tcolorbox}

\subsection{Ablation Study (RQ3)}

\begin{table*}[t]
\centering
\caption{Ablation study evaluating the performance contribution of each DeepSieve component. It shows how performance changes when components are removed ('w/o') or used in isolation. For each dataset, scores are presented in the format EM | F1 Score. }
\label{tab:ablation_study}
\begin{tabular}{lcccc|c|c}
\toprule
\textbf{Setting} & \textbf{Decomp.} & \textbf{Routing} & \textbf{Reflexion} & \textbf{HotpotQA} & \textbf{2Wiki.} & \textbf{MuSiQue} \\
\midrule
Full DeepSieve & \checkmark & \checkmark & \checkmark & 49.0 $\mid$ 61.6 & \textbf{62.8} $\mid$ \textbf{68.4} & \textbf{36.0} $\mid$ \textbf{46.8} \\
w/o Reflexion & \checkmark & \checkmark & & 17.3 $\mid$ 21.6 & 15.2 $\mid$ 15.4 & 5.4 $\mid$ 11.5 \\
w/o Routing & \checkmark & & \checkmark & \textbf{54.0} $\mid$ \textbf{65.8} & 61.4 $\mid$ 64.7 & 34.5 $\mid$ 44.6 \\
w/o Decomposition & & \checkmark & \checkmark & 33.1 $\mid$ 41.1 & 35.2 $\mid$ 36.9 & 22.0 $\mid$ 28.6 \\
Decomposition Only & \checkmark & & & 52.3 $\mid$ 62.1 & 60.3 $\mid$ 63.1 & 33.5 $\mid$ 44.0 \\
Routing Only & & \checkmark & & 32.1 $\mid$ 42.9 & 4.2 $\mid$ 4.6 & 6.4 $\mid$ 9.8 \\
Reflexion Only & & & \checkmark & 33.2 $\mid$ 43.0 & 28.7 $\mid$ 33.2 & 21.2 $\mid$ 28.1 \\
Naive RAG & & & & 30.4 $\mid$ 42.8 & 26.4 $\mid$ 31.3 & 20.5 $\mid$ 26.1 \\
\bottomrule
\end{tabular}
\end{table*}

\begin{figure}[t]
\centering
\includegraphics[width=0.85\linewidth]{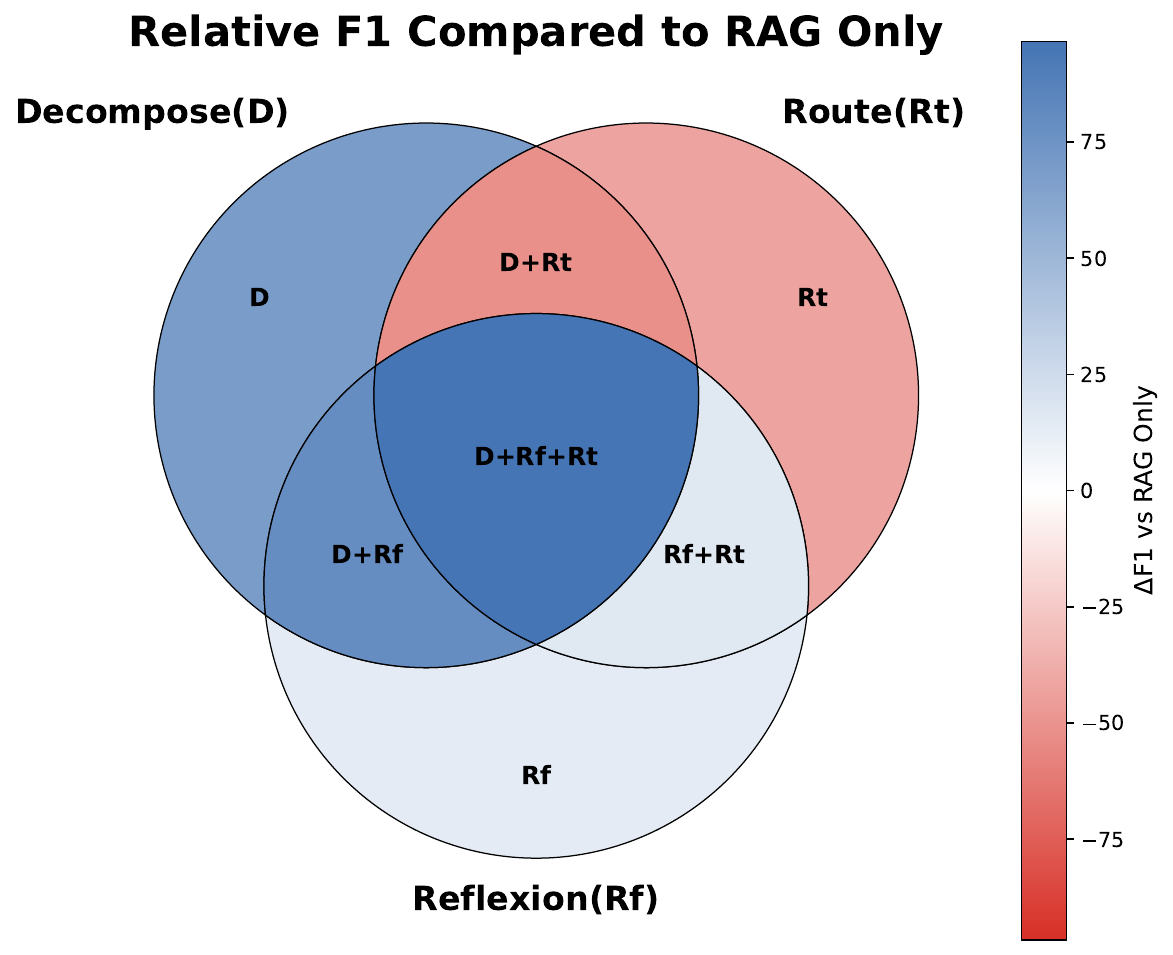}
\caption{Ablation study of F1 score improvements (blue) and declines (red) over Naive RAG across datasets. Color intensity corresponds to the magnitude of performance change, with darker shades indicating stronger effects.}
\label{fig:ablation_venn_plot}
\end{figure}

\begin{figure}[t]
\centering
\includegraphics[width=0.99\linewidth]{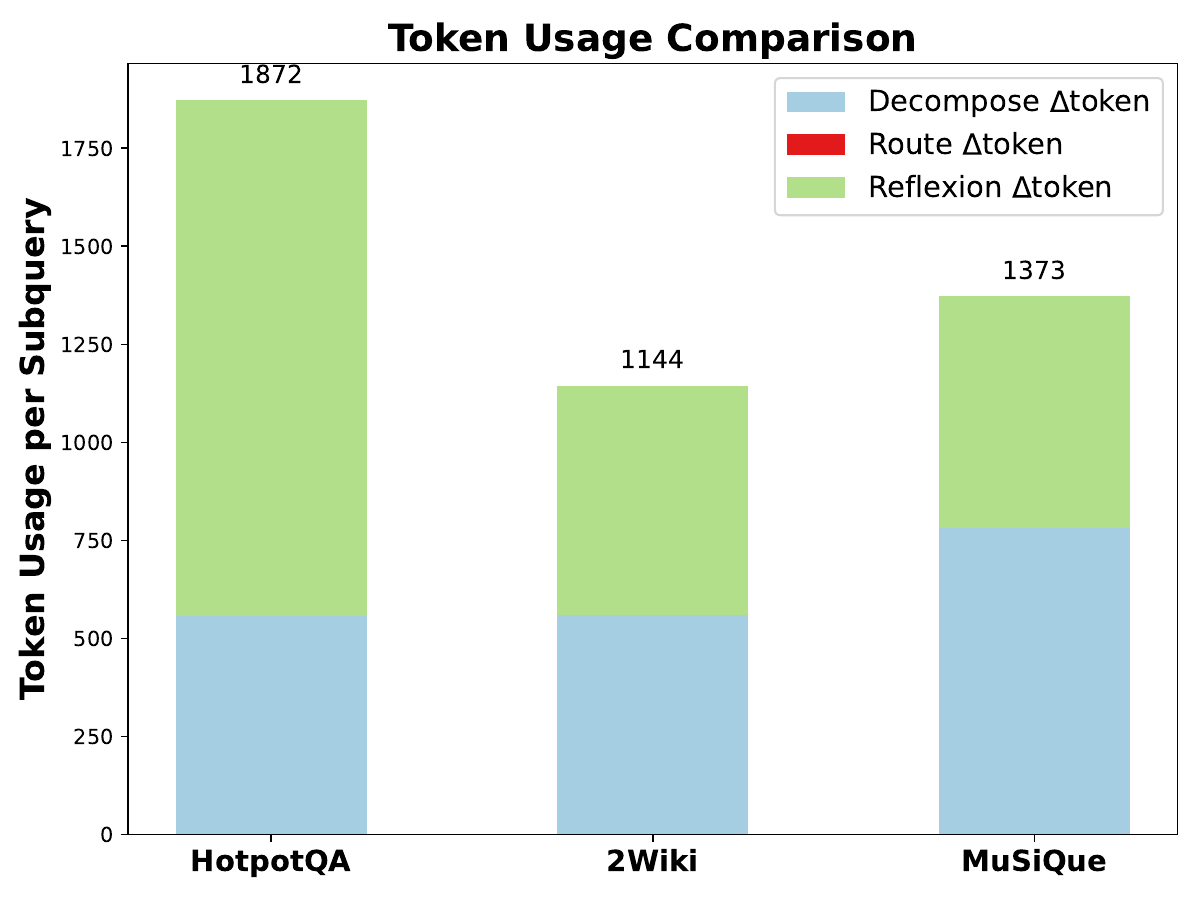}
\caption{A comparison of token costs for each stage of the framework. The stacked bars illustrate the cumulative token usage per subquery, showing the base cost of Decomposition (blue), plus the additional costs from Routing (red) and Reflexion (green).}
\label{fig:token_breakdown_plot}
\end{figure}
 
To understand the contribution of each module within DeepSieve, we conduct an ablation study to assess its individual and combined effects on system performance. Our results indicate that removing any module reduces performance, with \textbf{Reflexion} and \textbf{Decomposition} being most critical. Table~\ref{tab:ablation_study} shows that disabling Reflexion drops F1 from 68.4 to 15.4 on 2WikiMultihopQA, and removing Decomposition results in a 17.5-point drop on MuSiQue, highlighting their central roles in multi-hop reasoning.

In contrast, using \textbf{Routing} alone performs poorly and even slightly reduces accuracy on HotpotQA. This is likely because HotpotQA contains fewer source ambiguities, so the added complexity of Routing does not provide enough benefit. However, Figure~\ref{fig:ablation_venn_plot} shows that when Routing is combined with Decomposition and Reflexion (D+Rt+Rf), performance improves consistently across all datasets. While the D+Rf setup is already strong, adding Routing further boosts robustness and retrieval accuracy, especially in more diverse settings. Importantly, Routing is essential for DeepSieve’s ability to handle multi-source, heterogeneous knowledge. The fact that Routing matches prior performance without sacrificing accuracy in these more general scenarios demonstrates its effectiveness. Even in datasets without pronounced source heterogeneity, it generalizes well.

Other than performance, Figure~\ref{fig:token_breakdown_plot} details the token consumption of each module. The Decomposition stage consistently establishes a significant baseline cost, ranging from approximately 550 to 780 tokens per query. In contrast, the additional overhead from the Routing stage is negligible across all datasets. The most substantial portion of the additional cost is contributed by the Reflexion stage.
This cost analysis underscores the strategic design of our framework. With its minimal token footprint, the 
Routing module is the key component that enables DeepSieve to handle heterogeneous data sources and provides a net performance boost when combined with the other modules. Conversely, while Decomposition and Reflexion account for the majority of the token cost, the ablation results confirm they are indispensable for achieving high accuracy, as removing either leads to a dramatic drop in performance.

\begin{figure}[t]
    \centering
    \includegraphics[width=\linewidth]{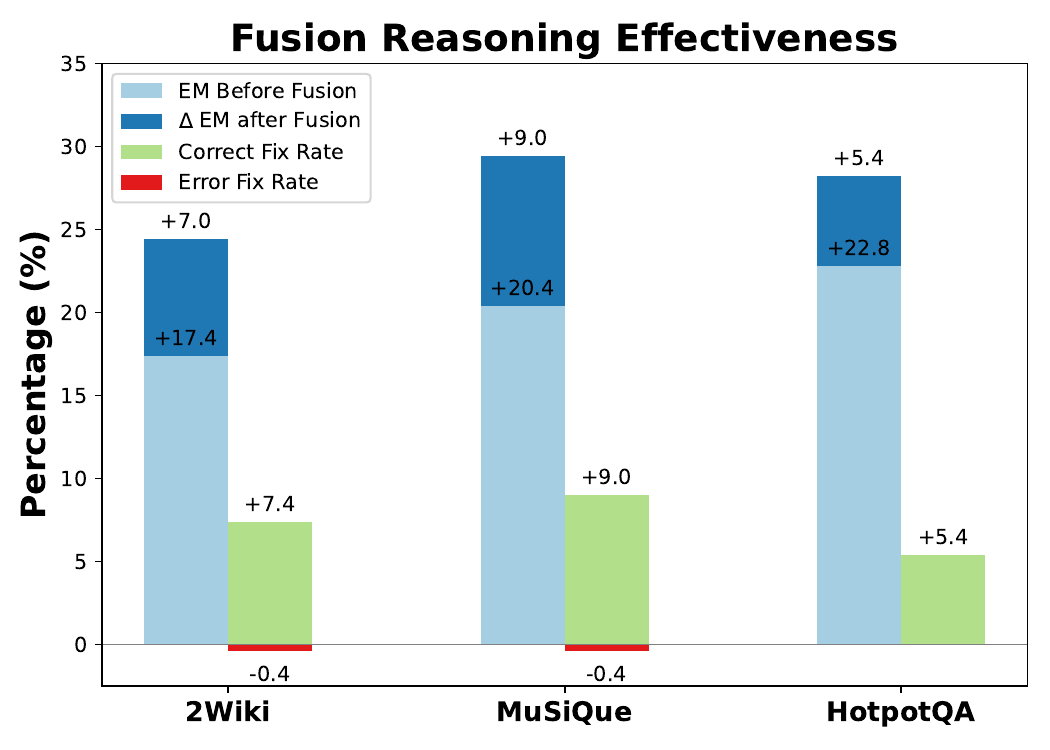}
    \caption{Effect of fusion reasoning on EM scores of DeepSieve(GraphRAG). For each dataset, the left bar shows EM before and after fusion (blue), and the right bar shows how often fusion fixes incorrect answers (green) or corrupts correct ones (red).}
    \label{fig:fusion_effectiveness}
    \vspace{-10pt}
\end{figure}

Finally, we evaluate the \textbf{Fusion} module's effectiveness by comparing fused outputs against final-subquery answers before fusion. Results demonstrate consistent accuracy improvements, with a high fix-to-error ratio confirming robust aggregation of routed subquery results as Figure~\ref{fig:fusion_effectiveness} shows.

\begin{tcolorbox}[
  colback=blue!5!white,
  colframe=black,
  coltitle=white,
  title=Takeaway: RQ3,
  fonttitle=\bfseries,
  sharp corners=southwest,
  enhanced,
  attach boxed title to top center={
    yshift=-2mm
  },
  boxed title style={
    colback=black,
    size=small,
    boxrule=0pt,
    rounded corners=southeast,
    sharp corners=north
  }
]
Decomposition and reflexion are key to accuracy. Routing alone may underperform, but in combination, it improves robustness and handles heterogeneous knowledge sources without sacrificing performance.
\end{tcolorbox}


\subsection{Modular Design and Adaptability (RQ4)}

DeepSieve supports both Naive RAG and GraphRAG retrieval setups, demonstrating its adaptability and modular design. To simulate heterogeneous corpora, we partition each dataset into \textit{local} and \textit{global} segments, enabling subquestion-level routing to different sources. DeepSieve achieves strong performance across both retrieval modes, outperforming prior RAG baselines while maintaining flexible source integration. Furthermore, to enable multi-format access, we implement modular interfaces in our framework(like SQL experiment results in Appendix~\ref{appendix:multi_src}), supporting potential integration with databases and APIs. 


\begin{tcolorbox}[
  colback=blue!5!white,
  colframe=black,
  coltitle=white,
  title=Takeaway: RQ4,
  fonttitle=\bfseries,
  sharp corners=southwest,
  enhanced,
  attach boxed title to top center={
    yshift=-2mm
  },
  boxed title style={
    colback=black,
    size=small,
    boxrule=0pt,
    rounded corners=southeast,
    sharp corners=north
  }
]
DeepSieve generalizes across retrievers and source configurations, with modular support for structured sources (e.g., SQL, JSON) implemented.
\end{tcolorbox}

\section{Related Work}

We position our work at the intersection of four key research directions: decomposition for multi-hop reasoning, RAG with heterogeneous sources, LLM-based routing, and reflexion RAG method.

\paragraph{Multi-Hop Reasoning and Decomposition}
Some researchers find that the reasoning step length and knowledge recall can contribute to multi-hop QA accuracy~\cite{jin2024impact, jin2024disentangling}. Multi-hop question answering (QA) requires breaking down complex queries into simpler subtasks.  
Decomposed Prompting proposes a modular planner–executor framework to tackle complex reasoning tasks \cite{khot2023decomposedpromptingmodularapproach}.  
ADaPT dynamically determines when to decompose using planner-based feedback loops \cite{prasad2024adaptasneededdecompositionplanning}.  
DISC improves scalability by dynamically decomposing inference steps with memory efficiency \cite{light2025discdynamicdecompositionimproves}.  
SealQA integrates decomposition and verification into search-augmented LMs \cite{cui2024sealqaraising}.  
Ye et al. formalize decomposition as a representation-quality check in RAG \cite{zeng2025knowledge}.  
These methods enhance reasoning depth and modular execution but do not address source- or tool-aware routing in heterogeneous environments.

\paragraph{RAG with Heterogeneous Sources}
RAG systems augment LLMs with both structured and unstructured knowledge \cite{lewis2020retrieval,MMQA}.  
HippoRAG introduces memory alongside structured retrieval \cite{hippo}.  
HippoRAG2 extends it with continual memory using clustering and profiling \cite{gutierrez2025ragmemorynonparametriccontinual}.  
InfuserKI enhances LLMs with knowledge graphs via infusing \cite{wang2024infuserki}.  
AutoSchemaKG automates schema induction for knowledge graph construction from web corpora \cite{bai2025autoschemakgautonomousknowledgegraph}.  
These approaches handle heterogeneous memory and structure, but still rely on flat retrieval indexes without per-subquestion navigation. Beyond retrieval, routing-based agent frameworks can explicitly decide \emph{which} knowledge/tool/agent to invoke at each step; for example, Sun propose fast--slow routing for visual agents to switch between lightweight and deliberative reasoning modules \cite{sun2025visual}.

\paragraph{LLM as Router for Source-Aware Retrieval}
Recent work explored by LLMs to control retrieval behavior, but often under homogeneous assumptions. Probing is popular in the LLM area~\cite{jin2024exploring}, Probing-RAG \cite{baek-etal-2025-probing} leverages LLMs' self-reflection to guide document selection, but operates over a single unified corpus. 
OmniRouter \cite{mei2025omnirouterbudgetperformancecontrollable} introduces cost-aware retrieval routing over sub-indices, assuming similar retrieval formats. 
Toolformer \cite{schick2023toolformer} fine-tunes LLMs to call APIs, yet does not support structured routing or modular tool orchestration. By contrast, DeepSieve treats routing as a decision step, matching each subquestion to a specific \emph{source profile} drawn from a heterogeneous set of corpora and tools (e.g., SQL, APIs, RAG corpora). 

\paragraph{Agentic Methods}
Agentic systems empower LLMs to reason, plan, and act over multi-step inference chains. Some agentic methods are also used for RAG tasks.
ReAct \cite{yao2023react} merges tool use and thought generation in a unified loop. 
ReWOO \cite{zhang2023rewoo} decouples retrieval from reasoning to reduce token cost. 
MA-RAG \cite{nguyen2025maragmultiagentretrievalaugmentedgeneration} introduces agentic collaboration via CoT-based subquerying. 
Recent works have also explored optimizing agentic RAG systems through caching mechanisms~\cite{lin2025cachemechanismagentrag} to reduce latency and cost.
In contrast, DeepSieve builds upon this line of work by introducing an explicit modular planner-router-retriever loop, enabling fine-grained reasoning over decomposed subquestions, recursive reflexion over failures, and heterogeneous tool dispatch.

\section{Conclusion}

We present DeepSieve, a RAG method that addresses the limitations of traditional RAG pipelines in handling compositional queries and heterogeneous knowledge sources. DeepSieve decomposes queries into subquestions, routes each to the most suitable source using an LLM-based router, and performs retrieval and reflexion before aggregating final answers.
Through extensive experiments on three multi-hop QA benchmarks, we demonstrate that DeepSieve achieves strong performance across different LLMs, outperforming both RAG baselines and agentic baselines. The method also generalizes across retrieval backends (Naive RAG vs. GraphRAG) and supports the interfaces for SQL and JSON sources in the source code.


\section*{Limitations}
While DeepSieve demonstrates strong performance in modular reasoning and heterogeneous source routing, there remain directions for further enhancement:
First, the current routing mechanism selects only a coarse-grained (tool, source) pair for each subquestion. This limits the system’s ability to leverage fine-grained configurations, such as tool-specific parameters (e.g., retrieval depth, temperature, API mode) or function-level APIs. Future work could extend the action space to support parameterized tool selection, enabling more adaptive behavior and cost-aware decisions at inference time \citep{mei2025litecuacomputermcpserver}.

Second, although DeepSieve supports modular integration of private sources, it treats all subquestions uniformly across users. In real-world settings, however, different users may have personalized knowledge graphs, access patterns, or preferences. A promising direction is to incorporate personalized routing and memory modules, allowing the LLM to learn user-specific retrieval paths, preferred sources, or task priors, thus enabling long-term adaptation and user-centric QA behavior.

Third, the interpretability of the agentic decision-making process is critical for user trust and system debugging. While DeepSieve effectively routes queries, the rationale behind specific routing choices can be opaque. Future work could incorporate agentic explainer frameworks, such as SAGE \citep{han2025sage}, RAGRouter-Bench \citep{wang2026ragrouter} to provide granular interpretations of the router's internal features and verify the logic behind its source selection and decomposition steps.

Furthermore, to further optimize routing precision, we plan to move beyond surface-level confidence scores. Recent findings on concept depth \citep{jin2025exploring} reveal how LLMs acquire knowledge across layers. Future iterations of DeepSieve could leverage these layer-wise signals to distinguish between well-mastered internal knowledge and hallucinations, triggering retrieval tools only when the model's internal concept depth is insufficient.
We believe these extensions will further enhance the controllability and personalization of RAG systems in both industrial and research applications.

Finally, our current evaluation primarily focuses on standard multi-hop question answering benchmarks. We have not yet explored the applicability of DeepSieve in more dynamic, interactive, or simulation-based environments. Future work could investigate how our routing mechanism adapts to complex scenarios such as computer-use agent \citep{mei2025rwomretrievalaugmentedworldmodel}, game prototyping \citep{li2025cardiverseharnessingllmsnovel, li2025archseekretrievingarchitecturalcase, LI2026106756} ,digital twin persona simulation \citep{du2025twinvoicemultidimensionalbenchmarkdigital, dusimvbg} and activity recognition \citep{li2020thuir}, where information retrieval must align with evolving states or specific character personas.

\section*{Acknowledgements}
We would like to thank Mingyu Jin for his contributions to this work.

\bibliography{references}

\appendix

\clearpage

\section{Case Study: Error Avoidance via Decomposition and Routing}
\label{sec:appendix}

To demonstrate the benefits of our routing-based, multi-stage RAG pipeline, we present several qualitative examples in which DeepSieve significantly outperforms a standard flat RAG baseline. While quantitative metrics provide a global summary, these case studies are crucial for illustrating the distinct types of reasoning and retrieval failures that arise in open-domain multi-hop QA—and how DeepSieve’s modular design mitigates them.

Each case highlights a representative failure mode of Pure RAG, such as hallucinating unsupported facts, making entity confusions due to semantic proximity, failing to perform multi-step inference, or being brittle to partial retrieval. In contrast, DeepSieve breaks down complex queries into structured subqueries, dynamically routes each subquery to the most relevant knowledge source, and employs reflexion to recover when intermediate reasoning fails.

The selected queries capture diverse challenges commonly encountered in real-world QA tasks:
\begin{itemize}
\item \textbf{Resolution of nested references} (Case 1): Answering requires tracing multiple entity links, e.g., identifying a person via another person's biography.
\item \textbf{Interpretation of entity-to-entity relationships} (Case 2): Understanding geographic containment and relational structure between entities.
\item \textbf{Robustness to early-stage retrieval errors} (Case 3): Overcoming noisy or misleading top-k passages by revising and retrying.
\item \textbf{Disambiguation in entity-rich settings} (Case 4): Navigating multiple overlapping or semantically similar names across corpora.
\item \textbf{Temporal linking across events} (Case 5): Connecting causally or chronologically related events that span different documents.
\end{itemize}

In all cases, we contrast DeepSieve with a Pure RAG system, which retrieves directly from a merged corpus without decomposition, source-specific routing, or any fallback mechanisms. Such flat pipelines often retrieve semantically similar but irrelevant contexts or hallucinate answers in the absence of explicit grounding.



\subsection*{Case 1: Decomposition Avoids Hallucination}
\textbf{Query:} Who is the husband of the woman who founded the Flying Doctors service in Nigeria?

\vspace{5pt}
\begin{tcolorbox}[colback=red!1, colframe=red!30!black, title=Pure RAG (Failure), sharp corners=south]
\textbf{Top-k Retrieved:}
\begin{itemize}[leftmargin=*]
  \item ``Flying Doctors Nigeria was established to provide air ambulance services across West Africa.'' 
  \item ``History of emergency medical care in Nigeria.'' 
  \item ``Private medical companies in Africa.'' 
\end{itemize}

\textbf{Answer:} Dr. Oluyemi \\
\textbf{Error:} No founder info; hallucinated name.
\end{tcolorbox}

\begin{center}
\textit{In contrast, DeepSieve decomposes the query into reasoning steps and retrieves targeted evidence from relevant sources.}
\end{center}

\begin{tcolorbox}[colback=green!1, colframe=green!40!black, title=DeepSieve (Success), sharp corners=south]
\textbf{Decomposition:}
\begin{itemize}[leftmargin=*]
  \item Q1: Who founded Flying Doctors? $\rightarrow$ \textbf{Dr. Ola Orekunrin}
  \item Q2: Who is her husband? $\rightarrow$ No public info
\end{itemize}

\textbf{Routing:} Q1 $\rightarrow$ Global, Q2 $\rightarrow$ Local \\
\textbf{Top-k Retrieved:}
\begin{itemize}[leftmargin=*]
  \item ``Dr. Ola Orekunrin founded the Flying Doctors Nigeria service.'' 
  \item ``Medical entrepreneurship in West Africa.'' 
\end{itemize}

\textbf{Final Answer:} No public record of her husband.
\end{tcolorbox}

In this case, Pure RAG fails because the question contains a nested structure that requires identifying the founder before inferring her marital status. Without decomposition, the model hallucinates a name (Dr. Oluyemi) not supported by retrieved content. DeepSieve correctly separates the query into steps and prevents hallucination by explicitly handling reference resolution.

\vspace{10pt}
\subsection*{Case 2: Routing Improves Corpus Precision}
\textbf{Query:} What country is the birthplace of Erik Hort a part of?

\vspace{5pt}
\begin{tcolorbox}[colback=red!1, colframe=red!30!black, title=Pure RAG (Failure), sharp corners=south]
\textbf{Top-k Retrieved:}
\begin{itemize}[leftmargin=*]
  \item ``Erik Hort was born in Montebello.'' 
  \item ``Cinderella attended the royal ball.'' 
\end{itemize}

\textbf{Answer:} United States \\
\textbf{Error:} Guesswork; no reasoning trace.
\end{tcolorbox}

\begin{center}
\textit{By decomposing the query and rerouting subquestions, DeepSieve uncovers geographical dependencies that Pure RAG missed.}
\end{center}

\begin{tcolorbox}[colback=green!1, colframe=green!40!black, title=DeepSieve (Success), sharp corners=south]
\textbf{Decomposition:}
\begin{itemize}[leftmargin=*]
  \item Q1: Who was born in Montebello? $\rightarrow$ Erik Hort
  \item Q2: What state is Montebello in? $\rightarrow$ New York
  \item Q3: What country is New York in? $\rightarrow$ United States
\end{itemize}

\textbf{Routing:} Local $\rightarrow$ Global $\rightarrow$ Global \\
\textbf{Top-k Retrieved:}
\begin{itemize}[leftmargin=*]
  \item ``Montebello is located in New York.'' 
  \item ``New York is one of the 50 US states.'' 
\end{itemize}

\textbf{Final Answer:} United States
\end{tcolorbox}

Here, a seemingly straightforward question actually requires multiple geographic hops. The Pure RAG system answers correctly by chance but lacks any justification. DeepSieve surfaces the latent structure, allowing the answer to be transparently derived through chain-of-thought and cross-source coordination.

\subsection*{Case 3: Reflexion Corrects Early Failure}
\textbf{Query:} What is one of the stars of “The Newcomers” known for?

\vspace{5pt}
\begin{tcolorbox}[colback=red!1, colframe=red!30!black, title=Pure RAG (Failure), sharp corners=south]
\textbf{Top-k Retrieved:}
\begin{itemize}[leftmargin=*]
  \item ``Chris Evans starred in ‘The Newcomers’.'' 
  \item ``Kate Bosworth acted in several teen films.''
\end{itemize}

\textbf{Answer:} Dano is an indie film actor. \\
\textbf{Error:} Off-target guess; no grounding.
\end{tcolorbox}

\begin{center}
\textit{DeepSieve’s reflexion mechanism re-queries when early steps fail, correcting misaligned or unsupported answers.}
\end{center}

\begin{tcolorbox}[colback=green!1, colframe=green!40!black, title=DeepSieve (Success via Reflexion), sharp corners=south]
\textbf{Decomposition:}
\begin{itemize}[leftmargin=*]
  \item Q1: Who starred in “The Newcomers”? $\rightarrow$ Chris Evans
  \item Q2: What is Chris Evans known for? $\rightarrow$ Captain America
\end{itemize}

\textbf{Routing:} Local $\rightarrow$ Global \\
\textbf{Top-k Retrieved:}
\begin{itemize}[leftmargin=*]
  \item ``Chris Evans played Captain America in the Marvel series.'' 
\end{itemize}

\textbf{Final Answer:} Captain America
\end{tcolorbox}

Reflexion plays a key role when intermediate retrieval fails. Here, the initial step is successful, but the answer to the second query is incorrect. DeepSieve detects this and re-attempts the step with an updated subquery. The correct answer emerges only after this re-routing, which is not possible in flat RAG systems.

\subsection*{Case 4: Routing Prevents Misleading Associations}
\textbf{Query:} Which Harry Potter character was played by Robbie Coltrane?

\vspace{5pt}
\begin{tcolorbox}[colback=red!1, colframe=red!30!black, title=Pure RAG (Failure), sharp corners=south]
\textbf{Top-k Retrieved:}
\begin{itemize}[leftmargin=*]
  \item ``Robbie Coltrane is a British actor.'' 
  \item ``He appeared in several comedies.'' 
\end{itemize}

\textbf{Answer:} Newt Scamander \\
\textbf{Error:} Wrong movie, wrong actor.
\end{tcolorbox}

\begin{center}
\textit{Naive RAG suffers from semantically similar distractions. DeepSieve routes to the most accurate corpus for disambiguation.}
\end{center}

\begin{tcolorbox}[colback=green!1, colframe=green!40!black, title=DeepSieve (Success), sharp corners=south]
\textbf{Decomposition:}
\begin{itemize}[leftmargin=*]
  \item Q1: Who is Robbie Coltrane? $\rightarrow$ Actor in Harry Potter
  \item Q2: What character did he play? $\rightarrow$ Rubeus Hagrid
\end{itemize}

\textbf{Routing:} Global $\rightarrow$ Global \\
\textbf{Top-k Retrieved:}
\begin{itemize}[leftmargin=*]
  \item ``Robbie Coltrane portrayed Hagrid in all eight Harry Potter films.'' 
\end{itemize}

\textbf{Final Answer:} Rubeus Hagrid
\end{tcolorbox}

This example shows how semantic similarity can be misleading. The name ``Robbie Coltrane'' triggers distractor documents about unrelated roles. DeepSieve’s global routing ensures selection of contextually relevant facts, avoiding confusion with similarly structured but irrelevant entries.

\subsection*{Case 5: Decomposition Enables Knowledge Linking}
\textbf{Query:} Who succeeded the Prime Minister that resigned during the Brexit vote?

\vspace{5pt}
\begin{tcolorbox}[colback=red!1, colframe=red!30!black, title=Pure RAG (Failure), sharp corners=south]
\textbf{Top-k Retrieved:}
\begin{itemize}[leftmargin=*]
  \item ``Brexit led to political unrest in the UK.'' 
\end{itemize}

\textbf{Answer:} Boris Johnson \\
\textbf{Error:} Lucky guess; no traceable chain.
\end{tcolorbox}

\begin{center}
\textit{Chained subquestions let DeepSieve explicitly link political timelines, uncovering correct facts.}
\end{center}

\begin{tcolorbox}[colback=green!1, colframe=green!40!black, title=DeepSieve (Success), sharp corners=south]
\textbf{Decomposition:}
\begin{itemize}[leftmargin=*]
  \item Q1: Who was UK PM during Brexit vote? $\rightarrow$ David Cameron
  \item Q2: Who succeeded David Cameron? $\rightarrow$ Theresa May
\end{itemize}

\textbf{Routing:} Global $\rightarrow$ Global \\
\textbf{Top-k Retrieved:}
\begin{itemize}[leftmargin=*]
  \item ``David Cameron resigned after Brexit.'' 
  \item ``Theresa May beat Cameron as PM.'' 
\end{itemize}

\textbf{Final Answer:} Theresa May
\end{tcolorbox}

This final case exemplifies how multi-hop question answering often entails chaining temporally or causally linked events. Without decomposition, the Pure RAG baseline simply guesses the answer based on limited evidence. DeepSieve reconstructs the chain of political succession through its decomposed subqueries, leading to a verifiable and accurate response.

\subsection*{Case 6: Decomposition Avoids Hallucination}
\textbf{Query:} Who is the husband of the woman who founded the Flying Doctors service in Nigeria?

\vspace{5pt}
\begin{tcolorbox}[colback=red!1, colframe=red!30!black, title=Pure RAG (Failure), sharp corners=south]
\textbf{Top-k Retrieved:}
\begin{itemize}[leftmargin=*]
  \item ``Flying Doctors Nigeria was established to provide air ambulance services across West Africa.'' 
  \item ``History of emergency medical care in Nigeria.'' 
  \item ``Private medical companies in Africa.'' 
\end{itemize}

\textbf{Answer:} Dr. Oluyemi \\
\textbf{Error:} No founder info; hallucinated name.
\end{tcolorbox}

\begin{center}
\textit{In contrast, DeepSieve decomposes the query into reasoning steps and retrieves targeted evidence from relevant sources.}
\end{center}

\begin{tcolorbox}[colback=green!1, colframe=green!40!black, title=DeepSieve (Success), sharp corners=south]
\textbf{Decomposition:}
\begin{itemize}[leftmargin=*]
  \item Q1: Who founded Flying Doctors? $\rightarrow$ \textbf{Dr. Ola Orekunrin}
  \item Q2: Who is her husband? $\rightarrow$ No public info
\end{itemize}

\textbf{Routing:} Q1 $\rightarrow$ Global, Q2 $\rightarrow$ Local \\
\textbf{Top-k Retrieved:}
\begin{itemize}[leftmargin=*]
  \item ``Dr. Ola Orekunrin founded the Flying Doctors Nigeria service.'' 
  \item ``Medical entrepreneurship in West Africa.'' 
\end{itemize}

\textbf{Final Answer:} No public record of her husband.
\end{tcolorbox}

In this case, Pure RAG fails because the question contains a nested structure that requires identifying the founder before inferring her marital status. Without decomposition, the model hallucinates a name (Dr. Oluyemi) not supported by retrieved content. DeepSieve correctly separates the query into steps and prevents hallucination by explicitly handling reference resolution.

\vspace{10pt}

\subsection*{Case 7: Multi-Source Reasoning with SQL and RAG}
\textbf{Query:} Which scientist born in the 19th century is known for discovering radioactivity, and what country did she live in during World War I?

\vspace{5pt}
\begin{tcolorbox}[colback=red!1, colframe=red!30!black, title=Pure RAG (Failure), sharp corners=south]
\textbf{Top-k Retrieved:}
\begin{itemize}[leftmargin=*]
  \item ``Radioactivity was discovered in the late 19th century.'' 
  \item ``Famous scientists in chemistry and physics.'' 
  \item ``Women in science and war.'' 
\end{itemize}

\textbf{Answer:} Lise Meitner \\
\textbf{Error:} Missed date constraint; incorrect person selected.
\end{tcolorbox}

\begin{center}
\textit{In contrast, DeepSieve uses SQL to filter candidates by birth year, then RAG to confirm contributions and residency.}
\end{center}

\begin{tcolorbox}[colback=green!1, colframe=green!40!black, title=DeepSieve (Success), sharp corners=south]
\textbf{Decomposition:}
\begin{itemize}[leftmargin=*]
  \item Q1: Which scientists were born in the 19th century? $\rightarrow$ [SQL: Marie Curie, others]
  \item Q2: Who among them discovered radioactivity? $\rightarrow$ \textbf{Marie Curie}
  \item Q3: Where did she live during World War I? $\rightarrow$ France
\end{itemize}

\textbf{Routing:} Q1 $\rightarrow$ SQL, Q2/Q3 $\rightarrow$ Global RAG \\
\textbf{Top-k Retrieved:}
\begin{itemize}[leftmargin=*]
  \item ``Marie Curie was born in 1867 and discovered radioactivity.'' 
  \item ``She lived in Paris and worked in France during World War I.''
\end{itemize}

\textbf{Final Answer:} Marie Curie, France.
\end{tcolorbox}

DeepSieve combines structured filtering and semantic reasoning, correctly narrowing down candidates using SQL and then using RAG to identify the scientific contribution and wartime location. Pure RAG fails to apply the 19th-century constraint and misattributes the discovery.

\section{Modular sources}

Table~\ref{tab:source_modules} summarizes the modular source modules available in DeepSieve. Each source is represented as a (Tool, Corpus) pair with an associated profile string. These profiles serve as natural language descriptors visible to the LLM router during routing decisions. The table illustrates the heterogeneity across source types (structured, unstructured, external APIs, and semi-structured logs), as well as their access modes (e.g., SQL, RAG, JSON, API). This modular abstraction enables plug-and-play source integration and facilitates fine-grained control over per-subquestion retrieval.

\section{Prompt Examples}
\label{sec:prompt_examples}

DeepSieve's effectiveness stems from carefully designed prompts that implement our core methodology. Each prompt type serves a distinct purpose in the reasoning pipeline while maintaining consistency with our framework's principles.

\subsection*{Decomposition Prompt}
\begin{tcolorbox}[
    colback=gray!5,
    colframe=black,
    title=Decomposition Prompt,
    coltitle=white,
    fonttitle=\bfseries,
    colbacktitle=gray!70,
    sharp corners=south,
    enhanced,
    boxrule=0.8pt,
    arc=2pt
]
You are a question planner. Decompose the complex question into a sequence of atomic questions that can be answered using a single fact each. 

Original Query: \texttt{Who succeeded the Prime Minister that resigned during the Brexit vote?}

Decomposed Subquestions:
1. Who was the UK Prime Minister during the Brexit vote?
2. Who succeeded that Prime Minister?

Only output the list of subquestions in order. Do not include explanations.
\end{tcolorbox}

\textbf{Design Rationale:} This prompt operationalizes our information sieving principle (Section~\ref{sec:decomposition}) by forcing atomic decomposition. The constrained format ensures each subquestion targets exactly one retrievable fact while preserving dependency relationships. The "no explanations" requirement minimizes token usage for downstream processing.

\subsection*{Routing Prompt}
\begin{tcolorbox}[
    colback=gray!5,
    colframe=black,
    title=Routing Prompt,
    coltitle=white,
    fonttitle=\bfseries,
    colbacktitle=gray!70,
    sharp corners=south,
    enhanced,
    boxrule=0.8pt,
    arc=2pt
]
You are a routing assistant. Your task is to decide whether a query should be answered using which tool-data pair.

Available Pairs:
- \texttt{local}: people and entity-specific information.
- \texttt{global}: general world knowledge including geography, history, etc.

Query: \texttt{What state is Montebello located in?}

Please output only one word: \texttt{local} or \texttt{global}. Do not explain your choice.
\end{tcolorbox}

\textbf{Design Rationale:} Implements our source-aware routing through extreme output simplification. The binary choice format minimizes token overhead while enforcing discrete source selection. This aligns perfectly with our tool-corpus abstraction and enables efficient routing decisions.

\subsection*{Reflexion Prompt}
\begin{tcolorbox}[
    colback=gray!5,
    colframe=black,
    title=Reflexion Prompt,
    coltitle=white,
    fonttitle=\bfseries,
    colbacktitle=gray!70,
    sharp corners=south,
    enhanced,
    boxrule=0.8pt,
    arc=2pt
]
You are a reflective reasoning agent. The previous attempt failed to find a valid answer. Try to rephrase or redirect the sub-question.

Failed Query: \texttt{What is one of the stars of "The Newcomers" known for?}  
Failed Result: \texttt{Dano is an indie film actor.} (not grounded)

Try to reflect and generate a new query that might work better.

Reflected Subquestion: \texttt{What is Chris Evans known for?}
\end{tcolorbox}

\textbf{Design Rationale:} Embodies our iterative refinement approach. The prompt structure guides the LLM to: (1) recognize retrieval failures, (2) preserve original intent, and (3) eliminate ambiguous references - all while maintaining the atomic question constraint.

\subsection*{Fusion Prompt}
\begin{tcolorbox}[
    colback=gray!5,
    colframe=black,
    title=Fusion Prompt,
    coltitle=white,
    fonttitle=\bfseries,
    colbacktitle=gray!70,
    sharp corners=south,
    enhanced,
    boxrule=0.8pt,
    arc=2pt
]
You are an answer synthesis agent. Given the original question and a list of subquestion-answer pairs, generate a final answer. Be concise and faithful to the evidence.

Original Question: \texttt{What country is the birthplace of Erik Hort a part of?}

Sub-QA Chain:
1. Who was born in Montebello? → Erik Hort  
2. What state is Montebello in? → New York  
3. What country is New York in? → United States

Final Answer: \texttt{United States}
\end{tcolorbox}

\textbf{Design Rationale:} Implements our evidence aggregation protocol (Section~\ref{sec:fusion}). The prompt enforces three key requirements: (1) traceability to subanswers, (2) conflict resolution, and (3) minimal elaboration - ensuring final outputs remain grounded in the retrieved evidence.

\section{Source Profiles for Routing}
\label{app:corpus_profiles}

To support per-subquery source routing, each corpus in DeepSieve is associated with a router-visible profile string that summarizes its scope and intended use. These profiles are automatically injected into the routing prompt (see Section~\ref{sec:routing}) to guide source selection. Below we list the local/global profile definitions used for each benchmark.

\subsection*{MuSiQue}

The MuSiQue benchmark utilizes two complementary knowledge sources with distinct characteristics:

\begin{tcolorbox}[
    colback=gray!5,
    colframe=black,
    title=Local Profile (Entity-Specific),
    coltitle=white,
    fonttitle=\bfseries,
    colbacktitle=gray!70,
    sharp corners=south,
    enhanced,
    boxrule=0.8pt,
    arc=2pt
]
This knowledge base focuses on specific named entities such as People (e.g., artists, politicians), Organizations (e.g., companies, institutions), Locations (e.g., cities, countries), Events (e.g., historical or sports events), and Works (e.g., books, films, artworks). It is ideal for entity-centric or biographical questions.
\end{tcolorbox}

\begin{tcolorbox}[
    colback=gray!5,
    colframe=black,
    title=Global Profile (Contextual Knowledge),
    coltitle=white,
    fonttitle=\bfseries,
    colbacktitle=gray!70,
    sharp corners=south,
    enhanced,
    boxrule=0.8pt,
    arc=2pt
]
This knowledge base contains more general background or contextual information that is not specific to any single named entity. It is more comprehensive and serves as a general fallback when the entity-specific source cannot answer the query. Select this profile when the query is not related to any specific entity in local profile.
\end{tcolorbox}

\subsection*{2WikiMultiHopQA}

For 2WikiMultiHopQA, we maintain the same local/global dichotomy but with adaptations for its unique multi-hop reasoning requirements:

\begin{tcolorbox}[
    colback=gray!5,
    colframe=black,
    title=Local Profile (Entity-Centric),
    coltitle=white,
    fonttitle=\bfseries,
    colbacktitle=gray!70,
    sharp corners=south,
    enhanced,
    boxrule=0.8pt,
    arc=2pt
]
This corpus includes documents related to People, Organizations, Locations, Events, and notable Works. It emphasizes named entities and is well-suited for questions involving specific persons, places, or creations.
\end{tcolorbox}

\begin{tcolorbox}[
    colback=gray!5,
    colframe=black,
    title=Global Profile (General Knowledge),
    coltitle=white,
    fonttitle=\bfseries,
    colbacktitle=gray!70,
    sharp corners=south,
    enhanced,
    boxrule=0.8pt,
    arc=2pt
]
This corpus provides general-purpose knowledge and covers a wide range of background facts. It should be used when queries are not focused on specific named entities or require general reasoning. Select this profile when the query is not related to any specific entity in local profile.
\end{tcolorbox}

\subsection*{HotpotQA}

HotpotQA's profiles are specially designed to handle its bridge and comparison questions, with clear separation between focused and diffuse knowledge:

\begin{tcolorbox}[
    colback=gray!5,
    colframe=black,
    title=Local Profile (Focused Facts),
    coltitle=white,
    fonttitle=\bfseries,
    colbacktitle=gray!70,
    sharp corners=south,
    enhanced,
    boxrule=0.8pt,
    arc=2pt
]
This local knowledge base contains concise factual descriptions about specific named entities. It includes individual biographies, locations, media works, historical figures, organizations, and structured encyclopedic entries. Each entry focuses on a single topic and offers concentrated factual coverage.

\textit{Examples of content:}
\begin{itemize}
  \item Biographies of notable people
  \item Descriptions of countries, cities, or institutions
  \item Synopses of TV shows, films, or novels
  \item Explanations of historical events or wars
\end{itemize}
\end{tcolorbox}

\begin{tcolorbox}[
    colback=gray!5,
    colframe=black,
    title=Global Profile (Diffuse Knowledge),
    coltitle=white,
    fonttitle=\bfseries,
    colbacktitle=gray!70,
    sharp corners=south,
    enhanced,
    boxrule=0.8pt,
    arc=2pt
]
This global knowledge base consists of diverse, loosely categorized facts and contextual information. It often includes edge cases, composite references, or entries that lack clear structural focus. These documents may span multiple entities, vague topics, or ambiguous scopes, making them harder to ground precisely.

\textit{Examples of content:}
\begin{itemize}
  \item Mentions of multiple entities without clear subject
  \item Abstract summaries or uncommon references
  \item Indirect relationships between people and places
\end{itemize}
\end{tcolorbox}

\section{Pseudocode Implementation}
\label{app:pseudocode}

The core logic is presented in two parts: the main control pipeline and the key LLM-driven helper functions. The main pipeline is designed as a highly modular and configurable controller. Its behavior is governed by a configuration object C, which allows for the systematic enabling or disabling of key components such as decomposition, routing, and reflexion.

\begin{tcolorbox}[
    title=Algorithm: The DeepSieve Pipeline,
    colback=gray!5,
    colframe=gray,
    fonttitle=\bfseries,
    coltitle=black,
    sharp corners=south,
    boxrule=0.5pt,
    arc=2pt,
    enhanced,
]
\textbf{Input:} Query $Q$, Set of source modules $S$, Config $C$ \\
\textbf{Output:} Final answer $\hat{A}$

\vspace{3pt}
$M \leftarrow \text{InitializeMemory()}$  \hfill \textcolor{gray}{// memory for results \& failures}

\If{$C.decompose$}{
    $\{q_i\} \leftarrow \text{Decompose}(Q)$ \hfill \textcolor{gray}{// Stage I: Decompose}
}
\Else{
    $\{q_i\} \leftarrow \{Q\}$ \hfill \textcolor{gray}{// RAG-only setting}
}

\For{each subquery $q_i$ in execution order}{
    $q_i^{actual} \leftarrow \text{SubstituteVariables}(q_i, M)$ \\
    $a_i \leftarrow \text{null}$; \quad $is\_success \leftarrow \text{false}$; \quad $j \leftarrow 0$ \\
    
    \While{$j < C.max\_reflexion\_attempts$ and not $is\_success$}{
        \If{$C.use\_routing$}{
            $s_i \leftarrow \text{Route}(q_i^{actual}, S, M.\text{failures})$ \hfill \textcolor{gray}{// Stage II: Route}
        }
        \Else{
            $s_i \leftarrow S_{merged}$ \hfill \textcolor{gray}{// RAG-only: use merged}
        }

        $a_{candidate} \leftarrow s_i.\text{Execute}(q_i^{actual})$ \hfill \textcolor{gray}{// retrieve or tool call} \\
        $a_i, is\_success \leftarrow \text{ExtractAnswer}(a_{candidate})$ \hfill \textcolor{gray}{// Stage III: Observe}

        \If{not $is\_success$ and $C.use\_reflexion$}{
            $M.\text{LogFailure}(q_i^{actual}, s_i)$ \hfill \textcolor{gray}{// reflect on failure}
        }
        $j \leftarrow j + 1$
    }

    $M.\text{LogSuccess}(q_i, a_i)$
}

$\hat{A} \leftarrow \text{Fuse}(Q, M)$ \hfill \textcolor{gray}{// Stage IV: Fuse all answers} \\
\Return $\hat{A}$
\end{tcolorbox}

This design is a significant advantage as it was crucial for conducting the ablation studies presented in our experiments. The algorithm's versatility is highlighted by the if C.use\_routing block, which shows how the framework can execute its primary logic for heterogeneous sources or gracefully degrade to a standard "RAG-only" baseline operating on a single merged source.

\begin{tcolorbox}[
    title=Algorithm: Helper Functions for DeepSieve,
    colback=gray!5,
    colframe=gray,
    fonttitle=\bfseries,
    coltitle=black,
    sharp corners=south,
    boxrule=0.5pt,
    arc=2pt,
    enhanced,
]
\textbf{Function} \texttt{Decompose}($Q$):\\
\quad $prompt \leftarrow \text{CreateDecompositionPrompt}(Q)$ \\
\quad $response \leftarrow \text{LLM}(prompt)$ \\
\quad \Return $\text{ParseSubqueryDAG}(response)$

\vspace{4pt}
\textbf{Function} \texttt{Route}($q$, $S$, $Fails$):\\
\quad $prompt \leftarrow \text{CreateRoutingPrompt}(q, S, Fails)$ \\
\quad $response \leftarrow \text{LLM}(prompt)$ \\
\quad \Return $\text{SelectSourceFrom}(response, S)$

\vspace{4pt}
\textbf{Function} \texttt{ExtractAnswer}($a_{candidate}$):\\
\quad $prompt \leftarrow \text{CreateExtractionPrompt}(a_{candidate})$ \\
\quad $response \leftarrow \text{LLM}(prompt)$ \\
\quad $(a, s) \leftarrow \text{ParseAnswerAndSuccess}(response)$ \\
\quad \Return $(a, s)$

\vspace{4pt}
\textbf{Function} \texttt{Fuse}($Q$, $M$):\\
\quad $prompt \leftarrow \text{CreateFusionPrompt}(Q, M.\text{successes})$ \\
\quad $response \leftarrow \text{LLM}(prompt)$ \\
\quad \Return $\text{ParseFinalAnswer}(response)$
\end{tcolorbox}

The reasoning process begins with an explicit planning step, where the Decompose function generates a structured Directed Acyclic Graph (DAG) of subqueries. A key strength of this approach is that it makes the reasoning plan transparent, machine-readable, and capable of modeling the complex, non-linear dependencies required to solve multi-hop questions. The execution of this plan is both dynamic and adaptive. For each subquery, the Route function intelligently selects a knowledge source, while the while loop embodies the Reflexion stage, providing a robust self-correction mechanism. If a retrieval is unsuccessful, the failure is logged to memory, allowing the system to attempt a different route on the next iteration.

This entire workflow is powered by the helper functions in the helper function algorithm, which abstract the core cognitive tasks into distinct, prompt-driven LLM calls. By encapsulating operations like Decompose and Fuse into modular functions, the framework adheres to the "LLM-as-a-Controller" paradigm. This has the advantage of making the system's logic transparent and easy to modify, as the behavior of each stage can be fine-tuned simply by refining its corresponding prompt without altering the core control flow code.


\section{Baseline Details}
\label{appendix:baseline_details}

We provide additional details for all baselines compared in our experiments.

\paragraph{ColBERTv2~\cite{santhanam2022colbertv2}.} A late-interaction dense retriever with efficient token-level matching. We use the open-sourced checkpoint trained on MS MARCO. It serves as the base retriever for IRCoT and HippoRAG.

\paragraph{IRCoT~\cite{gao2022interleaved}.} A strong multi-hop QA system that interleaves retrieval and CoT-style reasoning. We use their official implementation with ColBERTv2 as the retriever, following the original two-stage design.

\paragraph{HippoRAG~\cite{hippo}.} A neuro-inspired long-term memory RAG system that builds a hierarchical memory graph over retrieved content. We follow their open implementation and retrieval setup with ColBERTv2.

\paragraph{RAPTOR~\cite{nguyen2024raptor}.} A recent RAG framework using recursive abstraction and document-level graph indexing. We follow their standard OpenIE + KG construction pipeline and include their released knowledge graphs for fair comparison.

\paragraph{ReAct~\cite{yao2023react}.} An agent-style system that integrates reasoning and action via thought-action-observation loops. Our version uses the retrieval-augmented ReAct setup as implemented in their codebase.

\paragraph{ReWOO~\cite{zhang2023rewoo}.} An improved CoT method that trains workers and orchestrators to coordinate on reasoning and observation generation. We follow their implementation with the same planner-answerer split.

\paragraph{Reflexion~\cite{madaan2023selfrefine}.} A framework that iteratively refines answers based on self-evaluation. We implement a retrieval-augmented version where the model reflects after each retrieval failure.

\paragraph{Chain-of-Thought (CoT)~\cite{wei2022chain}.} A classic prompting baseline using few-shot reasoning examples without external retrieval. We use standard CoT prompts and apply them to each dataset individually.

\section{Dataset Descriptions}
\label{appendix:dataset}

We describe the three multi-hop QA datasets used in our experiments:

\paragraph{MuSiQue~\cite{trivedi2022musique}.} 
A challenging benchmark designed to test multi-hop and compositional reasoning. Each question requires aggregating facts across multiple Wikipedia passages. To reduce spurious correlations, MuSiQue provides both contrastive distractors and minimal context chains. We use the full MuSiQue-Answerable (MuSiQue-Full) version.

\paragraph{2WikiMultiHopQA~\cite{welbl2023multi}.} 
A clean and diverse multi-hop QA dataset built from Wikipedia entity pairs, where each question involves reasoning over two connected entities. It supports entity linking and is less noisy than HotpotQA. We follow prior work in using the dev split (1,000 samples) for evaluation.

\paragraph{HotpotQA~\cite{yang2018hotpotqa}.} 
A widely-used benchmark featuring bridge and comparison questions that require multi-hop reasoning over Wikipedia. Though popular, it is known to contain noise in both questions and support passages. We use the distractor setting, which includes 10 retrieved passages per question with only one or two relevant ones.

All datasets are preprocessed using IRCoT’s corpus construction pipeline, with supporting and distractor passages combined to form a flat retrieval index. The same set of questions is used across all methods for fair comparison.

\begin{table*}[t]
\centering
\caption{Modular source modules in DeepSieve and their router-visible profiles. Each source advertises a profile string that summarizes its domain and retrieval capability. The router relies on these profiles to select the most suitable source per subquestion.}
\label{tab:source_modules}
\resizebox{\textwidth}{!}{
\begin{tabular}{lll>{\raggedright\arraybackslash}p{9cm}}
\toprule
\textbf{Source Name} & \textbf{Type} & \textbf{Access Mode} & \textbf{Profile for Router Prompt} \\
\midrule
PersonnelDB   & Structured   & SQL         & Personnel records such as employee names, roles, and office locations. Use this source if the question involves internal or private company information. \\
Wikipedia     & Unstructured & RAG corpus  & This is a general-purpose encyclopedia covering broad world knowledge, including named entities, places, and public facts. \\
Google API    & External Tool& API         & This is a live web search API for retrieving real-time or geo-specific public information, such as current location, weather, or map-based data. \\
LogJSON       & Semi-structured & JSON     & This source holds semi-structured JSON logs from user chat or activity history. Use it when the query refers to personal history or past interactions. \\
\bottomrule
\end{tabular}
}
\end{table*}

\section{Experiment Setting Details}

To ensure consistency and reproducibility across all stages of reasoning, we adopt a unified LLM inference setup for all core components in DeepSieve—including decomposition, routing, reflexion, and final answer fusion. Each of these modules invokes a chat-based language model via HTTP API following the OpenAI-compatible \texttt{/chat/completions} protocol.

We use \textbf{deterministic decoding} by setting the generation temperature to 0.0, ensuring that the same input yields the same output across runs. This is critical for stability during iterative retrieval and reflection chains, and avoids stochastic drift in multi-hop inference.

By default, we use either \texttt{gpt-4o} or \texttt{deepseek-chat} as the backend model. These are selected based on experiment configuration or availability constraints. All prompts are constructed as single-turn queries, sent as a single \texttt{user}-role message without tool calling or system-level instructions. The response is expected in one pass, typically in JSON or structured text format.

\vspace{5pt}
\noindent\textbf{HTTP client settings.} All LLM requests are issued through a pooled session with adapter-level retries. The client is configured as follows:
\begin{itemize}
    \item \textbf{Temperature:} 0.0 (greedy decoding, no randomness)
    \item \textbf{Max retries:} 3
    \item \textbf{Timeout per request:} 60 seconds
    \item \textbf{Retry strategy:} Exponential backoff with delay $2^i$ for the $i$-th retry
    \item \textbf{Error conditions triggering retry:} HTTP status 429 (rate limit), 408 (timeout), 500–504 (server errors)
\end{itemize}

This robust client configuration minimizes disruptions during long-running experiments, where thousands of LLM queries may be issued in sequence. It also handles transient failures gracefully, such as momentary rate limits or backend latency spikes.

\vspace{5pt}
\noindent\textbf{Implementation details.} The entire LLM calling logic is encapsulated in a lightweight utility script:
\begin{quote}
\texttt{utils/llm\_call.py} implements \texttt{call\_openai\_chat(prompt, api\_key, model, base\_url)}, which handles inference, retry logic, and timeout handling in a self-contained function.
\end{quote}

This modular setup ensures that all DeepSieve modules, including planning, tool routing, and failure recovery, use a consistent and reliable interface to external language models. It also facilitates easy swapping between providers (e.g., OpenAI, DeepSeek, local vLLM) by simply changing the base URL and model string without modifying any internal pipeline logic.

\section{Mudular Experiment: Can Routing Preserve Expert Performance?}
\label{appendix:multi_src}

\paragraph{Motivation.}  
A key motivation for our work is the need to handle queries spanning heterogeneous knowledge sources—such as structured databases and unstructured corpora—that cannot be merged into a unified index. In real-world systems, queries may originate from diverse domains and require reasoning over non-coalescent sources like SQL tables and general-purpose document collections. To address this, DeepSieve introduces a routing module that decides, per query, which tool-data pair to invoke. But does this routing step degrade the performance of expert tools?

\vspace{4pt}
\paragraph{Setup.}
We design a diagnostic evaluation to test whether DeepSieve’s routing module preserves the upper-bound performance of expert tools when operating over distinct question types. We construct a test set consisting of two disjoint groups:

\begin{itemize}
  \item \textbf{Circle group}: factual questions about famous people (e.g., birth dates, professions), best answered via SQL queries.
  \item \textbf{Reg group}: general knowledge questions involving entities, events, and context, suitable for RAG-style text retrieval.
\end{itemize}

Each group is paired with a dedicated expert:
\begin{itemize}
  \item \textbf{Circle $\rightarrow$ SQLTool}: executes structured queries over a database of historical figures.
  \item \textbf{Reg $\rightarrow$ SimpleRAG}: answers queries using a standard vector retriever over text corpus.
\end{itemize}

\vspace{4pt}
\paragraph{Settings.}  
We compare two evaluation configurations:

\begin{itemize}
    \item \textbf{Setting 1 (Expert Oracle Routing)}: Each question is routed to its designated expert tool—Circle queries go to \texttt{SQLTool}, Reg queries go to \texttt{SimpleRAG}. This reflects the performance of a perfect oracle router.
    \item \textbf{Setting 2 (DeepSieve Routing)}: All queries are passed through DeepSieve’s LLM-based router, which must decide which tool to use without access to question type labels.
\end{itemize}

\vspace{4pt}
\paragraph{Results.}

\begin{table}[h]
\centering
\small
\setlength{\tabcolsep}{6pt}
\renewcommand{\arraystretch}{1.2}
\caption{DeepSieve vs. Expert Oracle performance (Exact Match $|$ F1). The underline means that the results are calculated in an ideal setting which is not feasible in real scene.}
\label{tab:routing-preserve}
\begin{tabular}{lccc}
\toprule
\textbf{Setting} & \textbf{SQL} & \textbf{RAG} & \textbf{Overall} \\
\midrule
Expert Oracle (S1) & 45.0 $|$ 46.2 & 32.4 $|$ 41.8 & \underline{38.7} $|$ \underline{44.0} \\
DeepSieve (S2) & 50.8 $|$ 52.0 & 48.3 $|$ 59.1 & 35.6 $|$ 40.1 \\
\bottomrule
\end{tabular}
\end{table}

\vspace{4pt}
\paragraph{Analysis.}
Setting 1 provides an upper bound under ideal conditions, where each question is perfectly matched with its corresponding expert tool. In contrast, Setting 2 reflects a realistic deployment scenario, where the system must decide the tool on the fly, without access to ground-truth labels or perfect routing hints.

Despite the performance drop in Setting 2 compared with the ideal setting, DeepSieve achieves reasonable performance in the absence of any oracle guidance. This validates the viability of routing-based reasoning under heterogeneous data regimes.

\vspace{4pt}
\paragraph{Why Routing is Necessary.}  
While Setting 1 performs better, it assumes prior knowledge of the best tool per question—an unrealistic expectation in practice. More importantly, real-world questions often span multiple sources, making joint reasoning across SQL and RAG data essential. Since these sources cannot be merged, a routing mechanism is indispensable.

This experiment demonstrates:
\begin{itemize}
    \item Routing is critical for modular, multi-source reasoning.
    \item DeepSieve can support this with acceptable trade-offs in performance.
    \item Even under idealized upper bounds, the performance gap is not catastrophic.
\end{itemize}

\section{Real-World Heterogeneous Experiment: MedQA + CaseHOLD}
\label{sec:appendix_real_world_exp}

\paragraph{Motivation.} 
While the experiments in former sections validate routing under controlled conditions, a key question still remains: can DeepSieve handle truly heterogeneous, real-world domains that are naturally incompatible? To address this, and to verify that our gains are not artifacts of artificial partitioning, we conducted an additional evaluation combining two distinct domain-specific corpora: \textbf{MedQA} (Medical) \cite{jin2021disease} and \textbf{CaseHOLD} (Legal) \cite{zhengguha2021}.

\paragraph{Setup.} 
We constructed a unified heterogeneous benchmark by mixing queries and corpora from MedQA and CaseHOLD. This setup mimics a realistic multi-tenant or multi-domain system where a single entry point must correctly route specialized queries to their respective knowledge bases (Medical vs. Legal) without cross-domain pollution.

\paragraph{Results.} 
Table~\ref{tab:medqa_casehold_res} presents the performance of DeepSieve against standard and Graph RAG baselines. DeepSieve significantly outperforms the baselines, achieving a +50\% improvement in Exact Match (EM) over Naive RAG.

\begin{table}[h]
    \centering
    \small
    \begin{tabular}{lcc}
        \toprule
        \textbf{Method} & \textbf{EM} & \textbf{F1} \\
        \midrule
        Naive RAG & 16.00 & 23.45 \\
        \textbf{DeepSieve (Naive RAG)} & \textbf{24.00} & \textbf{34.39} \\
        \midrule
        Graph RAG & 18.00 & 27.14 \\
        \textbf{DeepSieve (Graph RAG)} & \textbf{22.00} & \textbf{32.82} \\
        \bottomrule
    \end{tabular}
    \caption{Performance comparison on the combined MedQA (Medical) and CaseHOLD (Legal) setting. DeepSieve demonstrates superior adaptability to truly heterogeneous sources.}
    \label{tab:medqa_casehold_res}
\end{table}

\paragraph{Analysis.} 
The substantial performance gap highlights the critical role of source-aware routing in real-world scenarios. Standard RAG systems often suffer from "context pollution," retrieving irrelevant legal precedents for medical queries (or vice versa) due to vector space collisions. DeepSieve effectively isolates these domains via its routing stage.

\paragraph{Latency Note.} 
We also evaluated the wall-clock latency in this setting. DeepSieve averages \textbf{0.058s per query}, compared to 0.026s for Naive RAG. The slight overhead ($\sim$0.03s) incurred by the decomposition and routing steps is negligible for real-time applications, especially considering the significant gains in accuracy and the reduction in generated tokens.

\end{document}